\documentclass[9pt,twocolumn]{pnas-new}

\usepackage{echodefs}
\usepackage[utf8]{inputenc}
\usepackage[english]{babel}
\usepackage{bbm}
\usepackage{dsfont}
\usepackage{wrapfig}
\usepackage{booktabs}
\usepackage{graphicx}
\usepackage{subcaption}
\usepackage{microtype}
\usepackage{tikz}
\usepackage{amsmath}
\usepackage[normalem]{ulem}

\usepackage{xr}

\newtheorem{conjecture}{Conjecture}
\definecolor{myblue}{RGB}{42, 85, 127}
\definecolor{myred}{RGB}{239, 80, 118}
\definecolor{mygreen}{RGB}{70,189,156}
\definecolor{myyellow}{RGB}{250,205,110}
\definecolor{revgreen}{RGB}{20,150,20}

\def\?#1{}

\newcommand{\idd}[1]{{#1}}

\templatetype{pnasresearcharticle} 

\begin{document}

\title{Homophily modulates double descent generalization in graph convolution networks}

\author[a]{Cheng Shi}
\author[b,c,1]{Liming Pan}
\author[d]{Hong Hu}
\author[a,e,1]{Ivan Dokmani\'c}

\affil[a]{Departement Mathematik und Informatik, Universität Basel}
\affil[b]{School of Cyber Science and Technology, University of Science and Technology of China}
\affil[c]{School of Computer and Electronic Information, Nanjing Normal University}
\affil[d]{Wharton Department of Statistics and Data Science, University of Pennsylvania}
\affil[e]{Department of Electrical and Computer Engineering, University of Illinois at Urbana-Champaign}

\leadauthor{Shi}


\significancestatement{\idd{Graph neural networks (GNNs) have been applied with great success across science and engineering but we do not understand why they work so well. Motivated by experimental evidence of a rich phase diagram of generalization behaviors, we analyzed simple GNNs on a community graph model and derived precise expressions for generalization error as a function of noise in the graph, noise in the features, proportion of labeled data, and the nature of interactions in the graph. Computer experiments show that the analysis also qualitatively explains large ``production-scale'' networks and can thus be used to improve performance and guide hyperparameter tuning. This is significant both for the downstream science and for the theory of deep learning on graphs.}}

\authorcontributions{Author contributions: C.S., L.P., H.H. and I.D. designed research, provided mathematical analysis and wrote the paper; C.S. conducted experiments;}
\authordeclaration{The authors declare no competing interest.}
\correspondingauthor{\textsuperscript{1}To whom correspondence should be addressed. E-mail: panlm99@gmail.com, ivan.dokmanic@unibas.ch}

\keywords{Graph neural network $|$ transductive learning  $|$ double descent $|$ homophily $|$  stochastic block model}

\begin{abstract}
Graph neural networks (GNNs) excel in modeling relational data such as biological, social, and transportation networks, but the underpinnings of their success are not well understood. Traditional complexity measures from statistical learning theory fail to account for observed phenomena like the double descent or the impact of relational semantics on generalization error. Motivated by experimental observations of ``transductive'' double descent in key networks and datasets, we use analytical tools from statistical physics and random matrix theory to precisely characterize generalization in simple graph convolution networks on the contextual stochastic block model. Our results illuminate the nuances of learning on homophilic versus heterophilic data and predict double descent whose existence in GNNs has been questioned by recent work. We show how risk is shaped by the interplay between the graph noise, feature noise, and the number of training labels. Our findings apply beyond stylized models, capturing qualitative trends in real-world GNNs and datasets. As a case in point, we use our analytic insights to improve performance of state-of-the-art graph convolution networks on heterophilic datasets. 
\end{abstract}

\dates{This manuscript was compiled on \today}

\maketitle
\thispagestyle{firststyle}
\ifthenelse{\boolean{shortarticle}}{\ifthenelse{\boolean{singlecolumn}}{\abscontentformatted}{\abscontent}}{}

Graph neural networks (GNNs) recently achieved impressive results on problems as diverse as weather forecasting \cite{lam2022graphcast}, predicting forces in granular materials \cite{mandal2022robust}, or understanding biological molecules \cite{ingraham2019generative,gligorijevic2021structure,jumper2021highly}. They have become the de facto machine learning model for datasets with relational information such as interactions in protein graphs or friendships in a social network \cite{estrach2014spectral,defferrard2016convolutional,kipf2017semisupervised,hamilton2017inductive}. These remarkable successes triggered a wave of research on better, more expressive GNN architectures for diverse tasks, yet there is little theoretical work that studies why and how these networks achieve strong performance.

In this paper we study generalization in graph neural networks for transductive (semi-supervised) node classification: given a graph $G = (V, E)$, node features $\vx : V \to \mathbb{R}^F$, and labels $y : V_{\text{train}} \to \{-1, 1\}$ for a ``training'' subset of nodes $V_{\text{train}} \subset V$, we want to learn a rule which assigns labels to nodes in $V_{\text{test}} = V \setminus V_{\text{train}}$. This setting exhibits a richer generalization phenomenology than the usual supervised learning: in addition to the quality and dimensionality of features associated with data, the generalization error is affected by the quality of relational information (are there missing or spurious edges?), the proportion of observed labels $|V_{\text{train}}| / |V|$, and the specifics of interaction between the graph and the features. Additional complexity arises because links in different graphs encode qualitatively distinct semantics. Interactions between proteins are heterophilic; friendships in social networks are homophilic \cite{zhu2020beyond}. They result in graphs with different structural statistics, which in turn modulate interactions between the graphs and the features \cite{pei2020geom,chien2021adaptive}. Whether and how these factors influence learning and generalization is currently not understood. Outstanding questions include the role of overparameterization and the differences in performance on graphs with different levels of homophily or heterophily. Despite much work showing that in overparameterized models the traditional bias--variance tradeoff is replaced by the so-called double descent, there have been no reports nor analyses of double descent in transductive graph learning. Recent work speculates that this is due to implicit regularization \cite{Oono2020Graph}. 

Toward addressing this gap, we derive a precise characterization of  generalization in simple graph convolution networks (GCNs) in semi-supervised\footnote{More precisely, transductive.} node classification on random community graphs. We motivate this setting by first presenting a sequence of experimental observations that point to universal behaviors in a variety of GNNs on a variety of domains.

In particular, we argue that in the transductive setting a natural way to ``diagnose'' double descent is by varying the number of labels available for training (Section \ref{sec: motivation}). We then design experiments that show that double descent is in fact ubiquitous in GNNs: there is often a counterintuitive regime where more training data \emph{hurts} generalization \cite{nakkiran2021deep}. Understanding this regime has important implications for the (often costly) label collection and questions of observability of complex systems~\cite{liu2013observability}. While earlier work reports similar behavior in standard supervised learning, our transductive version demonstrates it directly \cite{nakkiran2021deep,chen2021multiple}. On the other hand, we indeed find that for many combinations of relational datasets and GNNs, double descent is mitigated by implicit or explicit regularization. Interestingly, the risk curves are affected not only by the properties of the models and data \cite{nakkiran2021deep}, but also by the level of homophily or heterophily in the graphs.

Motivated by these findings we then present our main theoretical result: a precise analysis of generalization on the contextual stochastic block model (CSBM) with a simple GCN. We combine tools from statistical physics and random matrix theory and derive generalization curves either in closed form or as solutions to tractable low-dimensional optimization problems. To carry out our theoretical analysis, we formulate a universality conjecture which states that in the limit of large graphs, the risks in GCNs with polynomial filters do not change if we replace random binary adjacency matrices with random Gaussian matrices. We empirically verify the validity of this conjecture in a variety of settings; we think it may serve as a starting point for future analyses of deep GNNs.

These theoretical results allow us to effectively explore a range of questions: for example, in Section \ref{sec: phenomena} we show that double descent also appears when we fix the (relative) number of observed labels, and vary relative model complexity (Fig.~\ref{fig: double descent with different alpha}). This setting is close but not identical to the usual supervised double descent \cite{belkin2020two}. We also explain why self-loops improve performance of GNNs on homophilic \cite{mcpherson2001birds} but not heterophilic \cite{pei2020geom,chien2021adaptive} graphs, as empirically established in a number of papers, but also that \emph{negative} self-loops benefit learning on heterophilic graphs \cite{wei2022understanding,baranwal2023optimality}. We then go back to experiment and show that building negative self-loop filters into state-of-the-art GCNs can further improve their performance on heterophilic graphs. This can be seen as a theoretical GCN counterpart of recent observations in the message passing literature \cite{wei2022understanding,baranwal2023optimality} and an explicit connection with heterophily for architectures such as GraphSAGE which can implement analogous logic \cite{hamilton2017inductive}.

Existing studies of generalization in graph neural networks rely on complexity measures like the VC-dimension or Rademacher complexity but they result in vacuous bounds which do not explain the observed new phenomena \cite{garg2020generalization,liao2021a,esser2021learning}. Further, they only indirectly address the interaction between the graph and the features. \idd{This interaction, however, is of key importance: an Erdős--Renyi graph is not likely to be of much use in learning with a graph neural network. In reality both the graph and the features contain information about the labels; learning should exploit the complementarity of these two views.}

\idd{Instead of applying the ``big hammers'' of statistical learning theory, we adopt a \emph{statistical mechanics} approach and study performance of simple graph convolution networks on the contextual stochastic block model (CSBM) \cite{deshpande2018contextual}. We derive precise expressions for the learning curves which exhibit a rich phenomenology.

The two ways to think about generalization, statistical learning theory and statistical mechanics, have been contrasted already in the late 1980s and the early 1990s. Statistical mechanics of learning, developed at that time by Gardner, Opper, Sejnowski, Sompolinsky, Tishby, Vallet, Watkin, and many others---an excellent account is given in the review paper by Watkin, Rau, and Biehl \cite{watkin1993statistical}---must make more assumptions about the data and the space of admissible functions, but it gives results that are more precise and more readily applied to the practice of machine learning.

These dichotomies have been revisited recently in the context of deep learning and highly-overparameterized models by Martin and Mahoney \cite{martin2017rethinking}, in reaction to Zhang et al.'s thought provoking  ``Understanding deep learning requires rethinking generalization'' \cite{DBLP:conf/iclr/ZhangBHRV17} which shows, among other things, that modern deep neural networks easily fit completely random labels. Martin and Mahoney explain that such seemingly surprising new behaviors can be effectively understood within the statistical mechanics paradigm by identifying the right order parameters and related phase diagrams. We explore these connections further in Section \ref{sec: discussion}---Discussion.}

\subsection*{Outline}

We begin by describing the motivational experimental findings in Section \ref{sec: motivation}. We identify the key trends to explain, such as the dependence of double descent generalization on the level of noise in features and graphs. In Section \ref{sec: precise analysis} we introduce our analytical model: a simple GCN on the contextual stochastic block model. Section \ref{sec: phenomena} then explores the implications of some of the analytical findings about self-loops and heterophily on the design of state-of-the-art GCNs. We follow this by a discussion of our results in the context of related work in Section \ref{sec: discussion}. In Section \ref{sec: main techniques} we explain the analogies between GCNs and spin glasses which allow us to apply analysis methods from statistical physics and random matrix theory. We follow with a few concluding comments in Section \ref{sec:conclusion}.

\section{Motivation: empirical results}\label{sec: motivation}

Given an $N$-vertex graph $G = (V, E)$ with an adjacency matrix $\mA \in \set{0, 1}^{N \times N}$ and features $\mX \in \R^{N \times F}$, a node classification GNN is a function $( \mA,\mX) \mapsto \vh({\vw}; \mA,\mX)$ insensitive to vertex ordering: for any node permutation $\pi$, $\vh({\vw}; \pi \mA \pi^\intercal,\pi \mX) = \pi \vh({\vw}; \mA,\mX)$. We are interested in the behavior of train and test risk,  
\begin{equation}\label{eqn: risk_N}
    R_N(S)=\frac{1}{|S|} \sum_{i\in S} \ell\left(\vy_i, \vh_{i}(\vw^*;\mA,\mX)\right),
\end{equation}
with $S \in \{V_\text{train}, V_\text{test}\}$ and $\ell(\cdot,\cdot)$ a loss metric such as the mean-squared error (MSE) or the cross-entropy. The optimal network parameters $\vw^*$ are obtained by minimizing the regularized loss 
\begin{equation}\label{eqn: loss_N}
L_N(\vw)= \frac{1}{|V_{\text{train}}|} \sum_{i\in V_{\text{train}}}  
\ell\left(\vy_i, 
    \vh_{i}(\vw;\mA,\mX) \right) + r_N(\vw),
\end{equation}
where $r_N(\vw)$ is a regularizer. 

\subsection*{Is double descent absent in GNNs?\?}


\begin{figure}[t]
        \centering 
        \includegraphics[width=\linewidth]{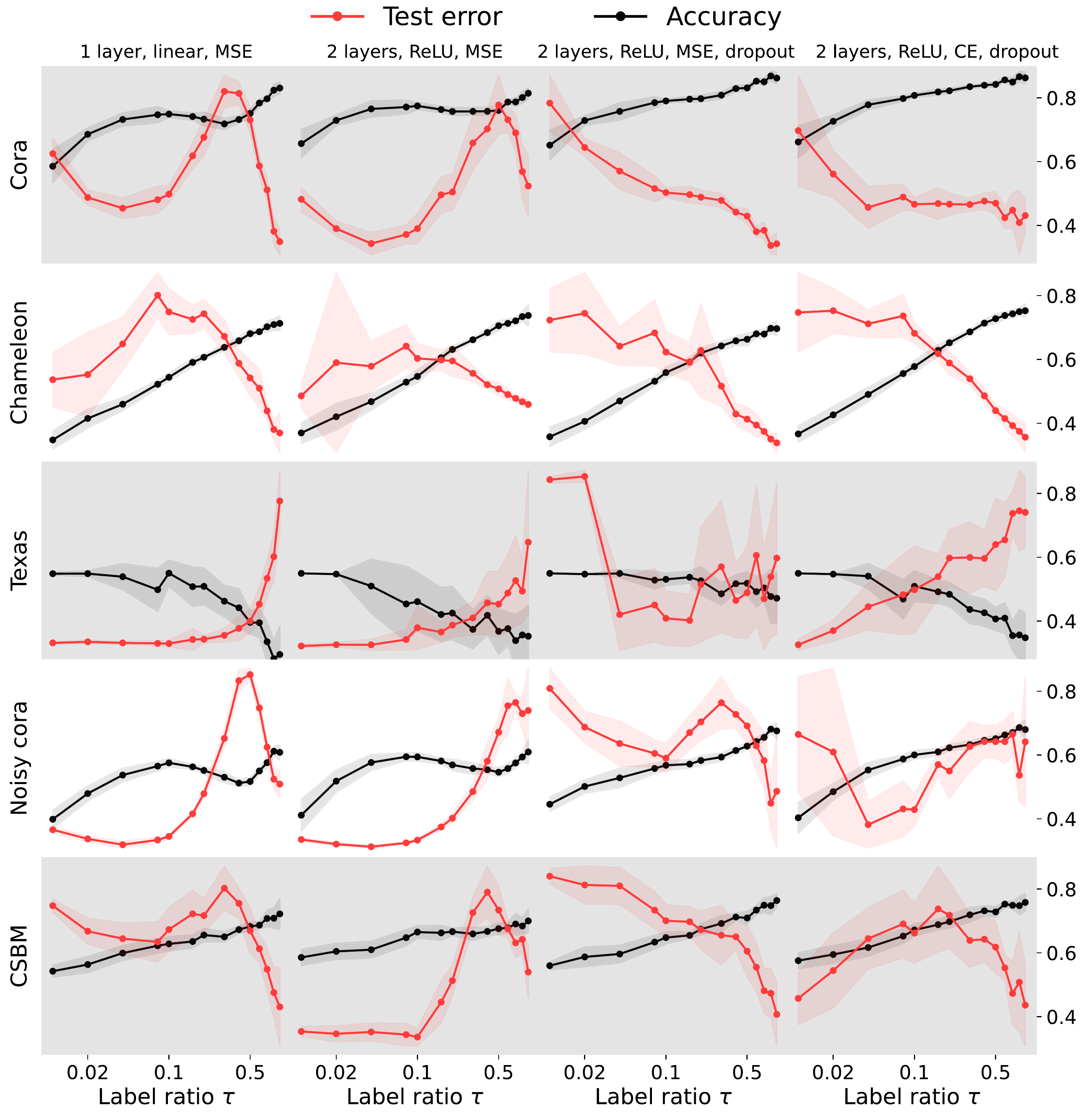}
        \caption{\idd{Double descent generalization for different GNNs, different losses, with and without explicit regularization, on datasets with varying levels of noise. We plot both the test error (\textcolor{red}{\textbf{red}}) and the test accuracy (\textbf{black}) against different training label ratios $\tau$ on the abscissa on a logarithmic scale. \textbf{First column}: one linear layer  trained by MSE loss; \textbf{second column}: a two-layer GCN with ReLU activations and MSE loss; \textbf{third column}: a two-layer GCN with ReLU activation function, dropout and MSE loss; \textbf{fourth column}: a two-layer GCN  with ReLU activations, dropout and cross-entropy loss;  Each experimental data point is averaged over 10 random train--test splits; the shadow area represents the standard deviation.
        The right ordinate axis shows classification accuracy; we suppress the left-axis ticks due to different numerical ranges.
        We observe that double descent is ubiquitous across datasets and architectures when varying the ratio of training labels: there often exists a regime where more labels impair generalization. }
        }
        \label{fig: Real DD}
\end{figure}

\idd{We start by investigating the lack of reports of \emph{double descent} in transductive learning on graphs. Double descent is a scaling of test risk with model complexity which is rather different from the textbook bias--variance tradeoff \cite{hastie2009elements,chen2021multiple}. Up to the interpolation point, where the model has sufficient capacity to fit the training data without error, things behave as usual, with the test risk first decreasing together with the bias and then increasing with the variance due to overfitting. But increasing complexity beyond the interpolation point---into an overparameterized region characteristic for modern deep learning---may make the test risk decrease again.

This generalization behavior has been identified already in the 90s by applying analytical tools from statistical mechanics to problems of machine learning; see for example Figure 10 in the paper by Watkin, Rau, and Biehl \cite{watkin1993statistical} or Figure 1 in Opper et al. \cite{opper1990ability} which show the generalization ability of the so-called pseudoinverse algorithm to train a boolean linear classifier (see also the book \cite{Engel_Van_den_Broeck_2001}). It is \emph{implicit} in work on phase diagrams of generalization akin to those for magnetism or the Sherrington--Kirkpatrick model \cite{seung1992statistical,opper1994learning}.

While these works are indeed the first to observe double descent, its significance for modern machine learning has been recognized by a line of research starting with \cite{belkin2019reconciling}. Double descent has been observed in complex deep neural networks \cite{nakkiran2021deep} and theoretically analyzed for a number of machine learning models \cite{watkin1993statistical,Engel_Van_den_Broeck_2001,belkin2020two,liao2020random,canatar2021spectral}. There are, however, scarcely any reports of double descent in graph neural networks. Oono and Suzuki \cite{Oono2020Graph} speculate that this may be due to implicit regularization in relation to the so-called oversmoothing \cite{li2018deeper}.}

\subsection*{Generalization in supervised vs transductive learning}

When illustrating double descent the test error is usually plotted against model complexity. For this to make sense, the amount of training data must be fixed, so the complexity on the abscissa is really \emph{relative complexity}; denoting the size of the dataset (node of nodes) by $N$ and the number of parameters by $F$ we let this relative complexity be $\alpha := F/N$. An alternative is to plot the risk against $\gamma = \alpha^{-1}$: Starting from a small amount of data (small $\gamma$), we first go through a regime in which increasing the amount of training data leads to \emph{worse} performance. In our context this can be interpreted as varying the size of the graph while keeping the number of features fixed. 

In transductive node classification we always observe the entire graph $\mA$ and the features associated with all vertices $\mX$, but only a subset of $M$ labels. It is then more natural to vary $\tau := M/N$ than $\alpha^{-1}$, with $M$ being the number of observed labels. \idd{Although the resulting curves are slightly different, they both exhibit double descent; in the terminology of Martin and Mahoney, both $\tau$ and $\alpha^{-1}$ may be called load-like parameters \cite{martin2017rethinking}; see also \cite{yang2021taxonomizing}.}\footnote{It may be interesting to note that papers by th physicists from the 90s put the amount of data on the abscissa \cite{watkin1993statistical,opper1990ability}.} In particular, they both have the interpolation peak at $\tau = \alpha^{-1}$, or $M = F$, when the system matrix becomes square and poorly conditioned. The key aspect of double descent is that the generalization error decreases on both sides of the interpolation peak.

Using $\tau$ instead of $\alpha^{-1}$ is convenient for several reasons: in real datasets, the number of input features is fixed; we cannot vary it. Further, there is no unique way to increase the number of parameters in a GNN and different GNNs are parameterized differently which complicates comparisons. Varying depth may lead to confounding effencts such as oversmoothing which is implicit regularization. Varying $\tau$ is a straightforward and clean way to compare different architectures in analogous settings. We can, however, easily vary $\alpha = \gamma^{-1}$ in our analytic model described in Section \ref{sec: precise analysis}; we show the related results in Fig.~\ref{fig: double descent with different alpha}.

\subsection*{Experimental observation of double descent in GNNs}


Armed with this understanding, we design an experiment as follows: we study the homophilic citation graph \texttt{Cora} \cite{sen2008collective} and the heterophilic graphs of Wikipeda pages \texttt{Chameleon} \cite{rozemberczki2021multi} and university web pages \texttt{Texas} \cite{pei2020geom}. We apply different graph convolution networks with different losses, with and without dropout regularization.  

Results are shown in Fig.~\ref{fig: Real DD}. Importantly, we plot both the test error (red) and the test accuracy (black) in node classification against a range of training label ratios $\tau$. In the first column, we use a one-layer GCN similar to the one we analyze theoretically in Section \ref{sec: precise analysis}, but with added degree normalization, self-loops, and multiple classes; in the second column, we use a two-layer GCN; in the third column we add dropout; in the fourth, we use the cross-entropy loss instead of the MSE.
This last model is used in the \texttt{pytorch-geometric} node classification tutorial\footnote{\href{https://pytorch-geometric.readthedocs.io/en/latest/notes/colabs.html}{https://pytorch-geometric.readthedocs.io/en/latest/notes/colabs.html}}.

First, with a one-layer network one can clearly observe transductive double descent on \texttt{Cora} in both the test risk and accuracy. The situation is markedly different on the heterophilic \texttt{Texas}, which contains only 183 nodes but 1703 features per node which yields relative model complexity $\alpha=F/N$ much higher than for other datasets. Here the test accuracy decreases near-monotonically, consistently with our theoretical analysis in Section \ref{sec: precise analysis} (cf. Fig.~\ref{fig_acc}\textbf{\textsf{D}}). In this setting strong regularization improves performance.

With a two-layer network the double descent still ``survives'' in the test error on \texttt{Cora}, but the accuracy is almost monotonically increasing except on \texttt{Texas}. These results corroborate the intuition that dropout and non-linearity alleviate GNN overfitting on node classification, especially for large training label ratios.

\begin{figure}[t]
    \centering
    \includegraphics[width=\linewidth]{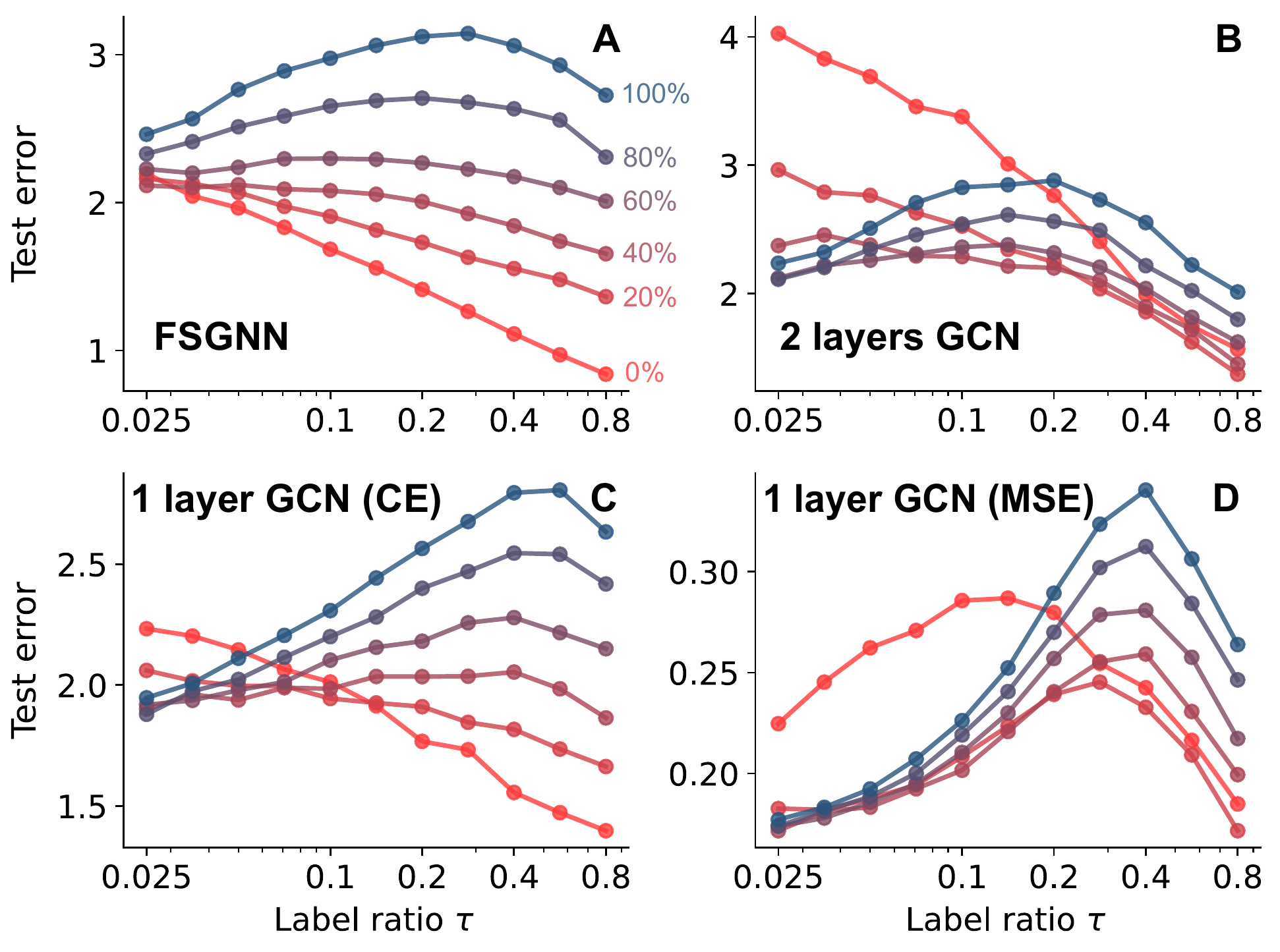}
    \caption{Test error with different training label ratios for different GCNs on \texttt{chameleon} (heterophilic) datasets. \textbf{\textsf{A}}: \texttt{FSGNN}\cite{maurya2021improving}; \textbf{\textsf{B}}: two- layer GCN with ReLU activations and cross-entropy loss; \textbf{\textsf{C}}: one layer GCN with cross entropy loss; (D): one layer GCN with MSE loss.   
    We interpolate between the original dataset shown in {\textcolor{myblue}{blue}}  ($0\%$ noise), and an Erdős–Rényi random graph shown in {\textcolor{red}{red}} ($100\%$ noise) by adding noise in increments of $20\%$. Noise is introduced by first randomly removing a given proportion of edges and then adding the same number of new random edges. The node features are kept the same. Each data point is averaged ten times, and the abscissa is on a logarithmic scale. We see that graph noise accentuates double descent, which is consistent with our theoretical results (see Fig.~\ref{fig_v}\textbf{\textsf{B}}). Similarly, better GNNs attenuate the effect where additional labels hurt generalization.}
    \label{fig:chameleon4_loss}
\end{figure}

We then explore the role of noise in the graph and in the features by manually adding noise to \texttt{Cora}. We randomly remove $30\%$ of the links and add the same number of random links, and randomize $30\%$ of the entries in $\mX$; results are shown in the fourth row of Fig.~\ref{fig: Real DD}. The double descent in test error appears even with substantial regularization. Comparing the first and the fourth row affirms that double descent is more prominent with noisy data; this is again consistent with our analysis (see Section \ref{sec: phenomena}).
In the last row we apply the networks to the synthetic CSBM. Observing the same qualitative behavior also in this case lends credence to the choice of CSBM for our precise analysis in Section \ref{sec: precise analysis}.

In Fig.~\ref{fig:chameleon4_loss} we further focus on the strongly heterophilic \texttt{Chameleon} which does not clearly show double descent in Fig.~\ref{fig: Real DD}. We randomly perturb different percentages of edges and in addition to GCNs also use the considerably more powerful FSGNN \cite{maurya2021improving}, which achieves current state-of-the-art results on \texttt{Chameleon}. Again, we see that double descent (a non-monotonic risk curve) emerges at higher noise (weaker heterophily). It is noteworthy that more expressive architectures do seem to mitigate double descent; conversely, a one-layer GCN exhibits double descent even without additional noise. We analytically characterize this phenomenon in Section \ref{sec: precise analysis} and illustrate it in Fig.~\ref{fig_v}. Beyond GCNs, we show that double descent occurs in more sophisticated GNNs like graph attention networks \cite{velivckovic2017graph}, GraphSAGE \cite{hamilton2017inductive} and Chebyshev graph networks \cite{defferrard2016convolutional}; see the SI Appendix \ref{APP: moreGCN} for details. 

In summary, transductive double descent occurs in a variety of graph neural networks applied to real-world data, with noise and implicit or explicit regularization being the key determinants of the shape of generalization curves. Understanding the behavior of generalization error as a function of the number of training labels is of great practical value given the difficulty of obtaining labels in many domains. For some datasets like \texttt{Texas}, using too many labels seems detrimental for some architectures. 

\section{A precise analysis of node classification on CSBM with a simple graph convolution network}\label{sec: precise analysis}

Motivated by the above discussions, we turn to a theoretical study of the performance of GCNs on random community graphs where we can understand the influence of all the involved parameters. We have seen in Section \ref{sec: motivation} that the generalization behavior in this setting qualitatively matches generalization on real data.

Graph convolution networks are composed of graph convolution filters and nonlinear activations. Removing the activations results in a so-called simple GCN \cite{wu2019simplifying} or a spectral GNN \cite{wang2022powerful,he2021bernnet}. For a graph $G = (V, E)$ with adjacency matrix $\mA$ and features that live on the nodes $\mX$,
\begin{equation}
    \label{eq:lin-gnc}
    \vh(\vw \mathop{;} \mA,\mX)=\mP(\mA) \mX \vw \quad \text{where} \quad \mP(\mA)=\sum_{k=0}^K c_k \mA^k,
\end{equation}
where $\vw \in \R^F$ are trainable parameters and $K$ is the filter support size in terms of hops on the graph. We treat the neighborhood weights $c_k$ at different hops as hyperparameters. We let $\mA^0 \bydef \mI_N$ so that the model \eqref{eq:lin-gnc} reduces to ordinary linear regression when $K=0$.


In standard feed-forward networks, removing the activations results in a linear end-to-end mapping. Surprisingly, GCNs without activations (such as SGC~\cite{wu2019simplifying}) or with activations only in the output (such as FSGNN~\cite{maurya2021improving} and GPRGNN~\cite{chien2021adaptive}) achieve state-of-the-art performance in many settings.\footnote{GCNs without activations are sometimes called ``linear'' in analogy with feed-forward networks, but that terminology is misleading. In graph learning, both $\mA$ and $\mX$ are bona fide parts of the input and a function which depends on their multiplication is a nonlinear function. What is more, in many applications $\mA$ is constructed deterministically from a dataset $\mX$, for example as a neighborhood graph, resulting in even stronger nonlinearity.}

We will derive test risk expressions for the above graph convolution network in two shallow cases: $\mP(\mA) = \mA$ and $\mP(\mA) = \mA + c\mI$. We will also state a universality conjecture for general polynomial filters. Starting with this conjecture, we can in principle extend the results to all polynomial filters using routine but tedious computation. We provide an example for the training error of a two-hop network in SI Appendix \ref{APP: RMT}. As we will show, this analytic behavior closely resembles the motivational empirical findings from Section \ref{sec: motivation}.

\subsection*{Training and generalization}

We are interested in the large graph limit $N \to \infty$ where the training label ratio $|V_{\text{train}}|  / N \to \tau$. We fit the model parameters $\vw$ by ridge regression $\vw^* := \argmin_{\vw}  L_{\mA,\mX}(\vw)$, where 
\begin{equation}\label{eqn: optim}
L_{\mA,\mX}(\vw)=\frac{1}{|V_{\text{train}}|}\sum_{i \in V_\text{train}} (\vy_i-\left(\vh_i(\vw;\mA,\mX)\right))^2+ \frac{r}{N} \|\vw\|^2_2.
\end{equation}
We are interested in the training and test risk in the limit of large graphs,
\begin{equation}\label{eqn: risk}
    R_{\mathrm{train}} = \lim_{N \to \infty} \mathbb{E} R_N(V_{\mathrm{train}}),
    \quad    R_{\mathrm{test}} = \lim_{N \to \infty} \mathbb{E} R_N(V_{\mathrm{test}}),
\end{equation}
as well as in the expected accuracy, 
\begin{equation}\label{eqn: ACC}
    \mathrm{ACC}
    =
    \lim_{N \to \infty} \mathbb{E} 
    \left[\frac{1}{|V_{\text{test}}|}
    \sum_{i\in V_\text{test}} \mathbf{1}\{ \vy_i=\mathrm{sign}(\vh_i\left(\vw^*\right))\} \right].
\end{equation}
We will sometimes write $R_\mathrm{train}(\mathcal{A})$, $R_\mathrm{test}(\mathcal{A})$, $\mathrm{ACC}(\mathcal{A})$ to emphasize that the matrix $\mA$ in \eqref{eq:lin-gnc} follows a distribution $\mathcal{A}$, $\mA \sim \mathcal{A}$. The expectations are over the random graph adjacency matrix $\mA$, random features $\mX$, and the uniformly random test--train partition $V = V_\mathrm{train} \cup V_\mathrm{test}$. Our analysis in fact shows that the quantities all concentrate around the mean for large $N$ (and $M$ and $F$): In the language of statistical physics, they are \emph{self-averaging}. \idd{This \emph{proportional aysmptotics} regime where $F, M$, and $N$ all grow large at constant ratios is more challenging to analyze than the regimes where dataset size or model complexity is constant, but it results in phenomena we see with production-scale machine learning models on real data; see also \cite{martin2017rethinking,liao2020random}.}

\subsection*{Contextual stochastic block model}

We apply the GCN to the contextual stochastic block model (CSBM). CSBM adds node features to the stochastic block model (SBM)---a random community graph model  \cite{deshpande2018contextual} where the probability of a link between nodes depends on their communities. The lower triangular part of the adjacency matrix $\mA^{\text{bs}}$ has distribution
\begin{equation}\label{eqn: A bs}
    \mathbb{P}\left( \mA_{ij}^{\text{bs}} = 1 \right)
    =
    \begin{cases}
        c_{\text{in}}/N & \text{ if } i \leq j \text{ and } \vy_{i} = \vy_{j} \\
        c_{\text{out}}/N & \text{ if } i \leq j \text{ and } \vy_{i}\neq \vy_{j}.
    \end{cases}
\end{equation}
A convenient parameterization is
\begin{equation*}
    c_{\text{in}} = d + \sqrt{d}\lambda, \quad \quad 
    c_{\text{out}} = d - \sqrt{d}\lambda,
\end{equation*}
where $d$ is the average node degree and the sign of $\lambda$ determines whether the graph is homophilic or heterophilic; $|\lambda|$ can be regarded as the graph signal noise ratio (SNR). 

We will also study a directed SBM \cite{wang1987stochastic,malliaros2013clustering} with adjacency matrix distributed as 
\begin{equation}\label{eqn: A bn}
    \mathbb{P}\left( \mA_{ij}^{\text{bn}} = 1 \right)
    =
    \begin{cases}
        c_{\text{in}}/N &  \quad \text{ if }  \vy_{i} = \vy_{j} \\
        c_{\text{out}}/N & \quad \text{ if } \vy_{i}\neq \vy_{j}.
    \end{cases}
\end{equation}
Many real graphs have directed links, including chemical connections between neurons, the electric grid, folowee--folower relation in social media, and Bayesian graphs. In our case the directed SBM facilitates analysis with self-loops while exhibiting the same qualitative behavior and phenomenology as the undirected one.

The features of CSBM follow the spiked covariance model,
\begin{equation}\label{eqn: scm}
    \vx^i = \sqrt{\frac{\mu}{N}}\vy_i \vu + \boldsymbol{\xi}^i,
\end{equation}
where $\vu \sim \mathcal{N}(0,\mI_F/F)$ is the $F$-dimensional hidden feature and $\boldsymbol{\xi}^i \sim \mathcal{N}(0,\mI_F/F)$ are i.i.d. Gaussian noise; the parameter $\mu$ is the feature SNR. We work in the proportional scaling regime where $\frac{N}{F} \to \gamma$, with $\gamma$ being the inverse relative model complexity, and ascribe feature vectors to the rows of the data matrix $\mX$,
\begin{equation}
    \mX = [\vx^1,\cdots,\vx^N]^\intercal =\sqrt{\frac{\mu}{N}}\vy \vu^\intercal  + \boldsymbol{\Xi}^x.
\end{equation}
We assume throughout that the two communities are balanced; without loss of generality we let $\vy_i = 1$ for $i = 1, 2, \ldots, {N}/{2}$ and $\vy_i = -1$ for $i > {N}/{2}$.

We will show that CSBM is a comparatively tractable statistical model to characterize generalization in GNNs. Intuitively, when $N \to \infty$, the risk should concentrate around a number that depends on five parameters:
\[\arraycolsep=10.4pt\def\arraystretch{1.3}
\begin{array}{cl}
\lambda & \text{Degree of homophily (Graph SNR)} \\
\mu       & \text{Feature SNR,} \\
\alpha   & \text{Relative model complexity } (=\gamma^{-1})\\
\tau      & \text{Label ratio} \\
r         & \text{Ridge regularization parameter.}
\end{array}
\]

We emphasize that we study the challenging weak-signal regime where $\lambda$, $\mu$ and $\gamma$ do not scale with $N$ (but $F$ does). This stands in contrast to recent machine learning work on CSBM \cite{lu2021learning, baranwal2021graph} which studies the low-noise regime where $\mu$ or $\lambda^2$ scale with $N$, or even
the noiseless regime where the classes become linearly separable after applying a graph filter or a GCN. We argue that the weak-signal regime is closer to real graph learning problems which are neither too easy (as in linearly separable) nor too hard (as with a vanishing signal). The fact that we discover phenomena which occur in state-of-the-art networks and real datasets supports this claim.

We outline our analysis in Section \ref{sec: main techniques} and provide the details in the SI appendices. But first, in the following section, we show that the derived expressions precisely characterize generalization of shallow GCNs on CSBM and also give a correct qualitative description of the behavior of ``big'' graph neural networks on complex datasets, pointing to interesting phenomena and interpretations. 


\section{Phenomenology of generalization in GCNs}\label{sec: phenomena}

We focus on the behavior of the test risk under various levels of graph homophily, emphasizing two main aspects: i) different levels of homophily lead to different types of double descent; ii) self-loops, standard in GCNs, create an imbalance between heterophilic and homophilic datasets; negative self-loops improve the handling of heterophilic datasets.

\subsection*{Double descent in shallow GCNs on CSBM}
\begin{figure*}[h]
        \centering
        \includegraphics[width=\textwidth]{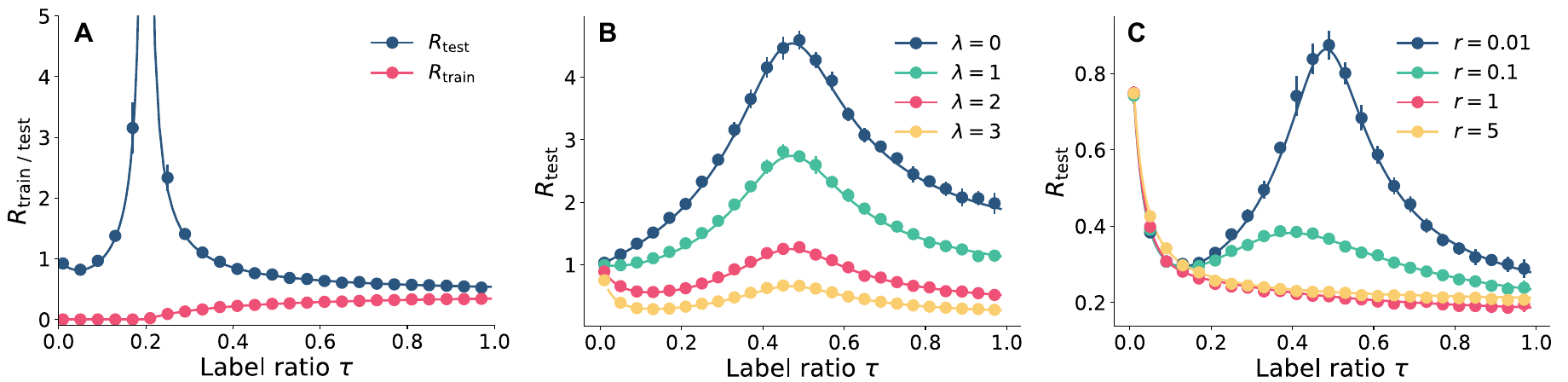}
        \caption{Theoretical results computed by the replica method (solid line) versus experimental results (solid circles) on CSBM, with $\mP(\mA)=\mA$, for varying training label ratios $\tau$. \textbf{\textsc{A}}: training and test risks with $\lambda=\mu=1$, $\gamma=5$ and $r=0$. (For $\tau<0.2$, we use the pseudoinverse in \eqref{eqn: w star} in numerics and $r=10^{-5}$ for the theoretical curves). We further study the impact of varying $\lambda$  in  \textbf{\textsf{B}} and $r$ in  \textbf{\textsf{C}}. We set $r=0.02$, $\gamma=2$, $\mu=1$ in  \textbf{\textsf{B}} and $\lambda=3$, $\mu=1$, $\gamma=2$ in  \textbf{\textsf{C}}. In all experiments we set $N=5000$ and $d=30$. We work with the symmetric binary adjacency matrix ensemble $\mathcal{A}^{\text{bs}}$. Each experimental data point is averaged over $10$ independent trials; their standard deviation is shown by vertical bars. The theoretical curves agree perfectly with experiments but also qualitatively with the phenomena we observed on real data in Section \ref{sec: motivation}.} 
        \label{fig_v}
\end{figure*}

As we show in Section \ref{sec: main techniques} and SI Appendix Section \ref{APP: Derivation Sketch}, the expression for the test risk for unregularized regression ($r = 0$) with shallow GCN can be obtained in closed form as
$$
    R_{\mathrm{test}} = \frac{\gamma\tau(\gamma+\mu)}{(\gamma\tau-1)\left(\gamma+\lambda^{2}(\mu+1)+\mu\right)}
$$
when $\gamma\tau>1$. It is evident that the denominator vanishes as $\gamma \tau$ approaches 1. When this happens, the system matrix $\mI_{\mathrm{train}}\mP(\mA)\mX$, where $\mI_{\mathrm{train}}$ selects the rows for which we have labels; see Section \ref{sec: main techniques}, \eqref{eqn: Itrain}, is square and near-singular for large $N$, which leads to the explosion of $R_{\mathrm{test}}$ (Fig.~\ref{fig_v}\textbf{\textsf{A}}).  When relative model complexity is high, i.e., $\gamma=N/F < 1$ is low, $\tau\gamma $ is always less than $1$. In such cases, no interpolation peak appears, which is consistent with our experimental results for the \texttt{Texas} dataset where $\gamma = 0.11$; cf. Fig.~\ref{fig_acc}\textbf{\textsf{D}}.

At the other extreme, for strongly regularized training (large $r$) the double descent disappears (Fig.~\ref{fig_v}\textbf{\textsf{C}}); it has been shown that this happens at optimal regularization \cite{mei2022generalization,canatar2021spectral}. The absolute risk values in Fig.~\ref{fig_v}\textbf{\textsf{B}} and \ref{fig_v}\textbf{\textsf{C}} show the same behavior. 

Fig.~\ref{fig_v}\textbf{\textsf{B}} shows that when the graph is very noisy ($\lambda$ is small)  the test error starts to increase as soon as the training label ratio $\tau$ increases from $0$. When $\lambda$ is large, meaning that the graph is discriminative, the test error first decreases and then increases. Similar behavior can be observed when varying the feature SNR $\mu$ instead of $\lambda$. Double descent also appears in test accuracy (Fig.~\ref{fig_acc}).

While these curves all illustrate double descent in the sense that they all have the interpolation peak on both sides of which the error decreases, they are qualitatively different. The emergence of these different shapes can be explained by looking at the distribution of the predicted $i$th label $\vh_i(\vw^*)$. As we show in SI Appendix \ref{APP: Derivation Sketch}, $\vh_i(\vw^*)$ is normally distributed with mean and variance given by the solutions of a saddle point equation outlined in Section \ref{sec: main techniques}. The test accuracy can thus be expressed by the error function (cf. \eqref{eqn: Risk r>=0}).

As we increase the number of labels, the mean $\mathbb{E}[\vh_i(\vw^*)]$ approaches $\vy_i$ monotonically. However, the variance $\mathrm{Var}[ \vh_i(\vw^*)]$ behaves diferently for different model complexities $\alpha = \frac{1}{\gamma}$ and regularizations $r$, resulting in distinct double descent curves.

For example, when $r \to 0$ and $\tau \to \frac{1}{\gamma}$, the variance of $\vh_{i}(\vw^*)$ for $i \in V_{\text{test}}$ diverges and the accuracy approaches $50\%$, a random guess. On the other hand, when $r$ is large, the variance is small and double descent is mild or absent, as shown in Fig.~\ref{fig_acc}\textbf{\textsf{A}}. Figure~\ref{fig_acc}\textbf{\textsf{B}} shows a typical double descent curve with two regimes where additional labels hurt generalization. In Fig.~\ref{fig_acc}\textbf{\textsf{C}} we also see a mild double descent when the relative model complexity is close to $1$: this is consistent with experimental observations on \texttt{Cora} in Fig.~\ref{fig: Real DD}. In certain extremal cases, for example when $\gamma$ is very small, the test accuracy continuously decreases after a very small ascent around $\tau=0$ (Fig.~\ref{fig_acc}\textbf{\textsf{D}}); this is consistent with our experimental observations for the \texttt{Texas} dataset.



\begin{figure}[!h]
    \centering
    \includegraphics[width=\linewidth]{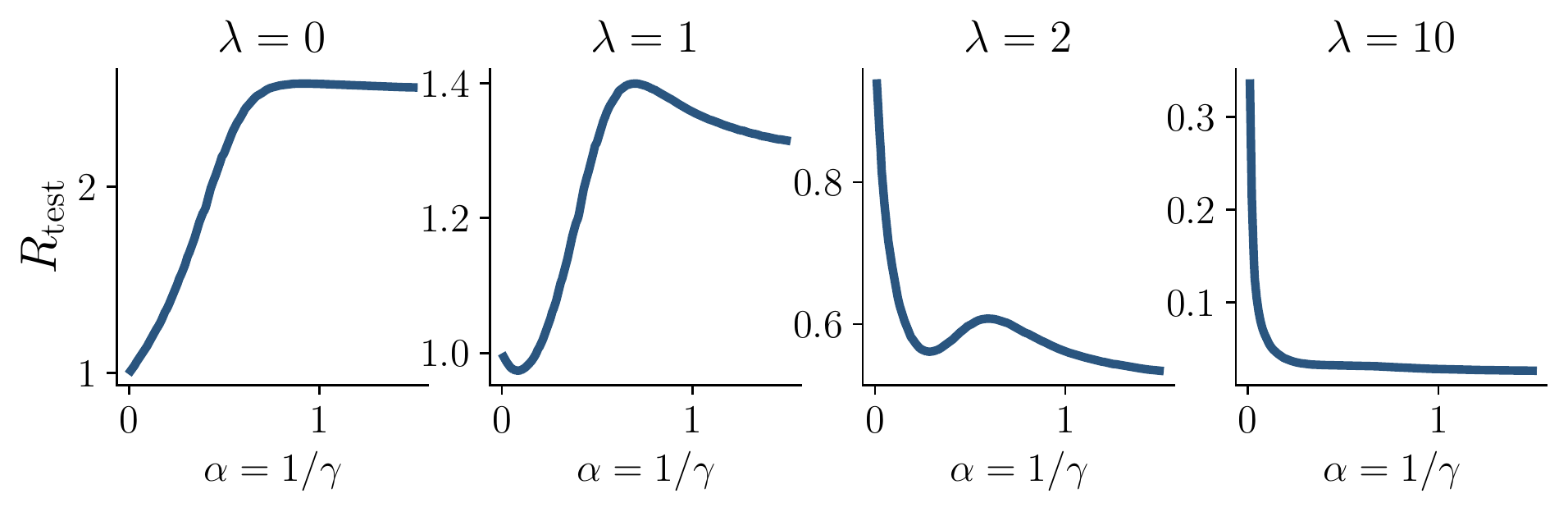}
    \caption{Test risk as a function of relative model complexity $\alpha = \gamma^{-1}$: different levels of homophily lead to distinct types of double descent in CSBM. Plots from left to right (with increasing $\lambda$) show curves for graphs of decreasing randomness. Varying model complexity in GNNs yields non-monotonic curves similar to those in the earlier studies of double descent studies in supervised (inductive) learning. Note that the overall shape of the curve is strongly modulated by the degree of homophily in the graph.}
    \label{fig: double descent with different alpha}
\end{figure}

\subsection*{Double descent as a function of the relative model complexity} As mentioned earlier, the theoretical model makes it easy to study double descent as we vary the model complexity $\alpha = 1 / \gamma$ rather than $\tau$; this is closer to the traditional reports of double descent in supervised learning. The resulting plots follow a similar logic: as shown in Fig.~\ref{fig: double descent with different alpha}, adding randomness in the graph (low $|\lambda|$), makes the double descent more prominent. Conversely, for a highly homophilic graph (large $\lambda$), the test risk decreases monotonically as the relative model complexity $\alpha$ grows.

\begin{figure}[t]
    \includegraphics[width=\linewidth]{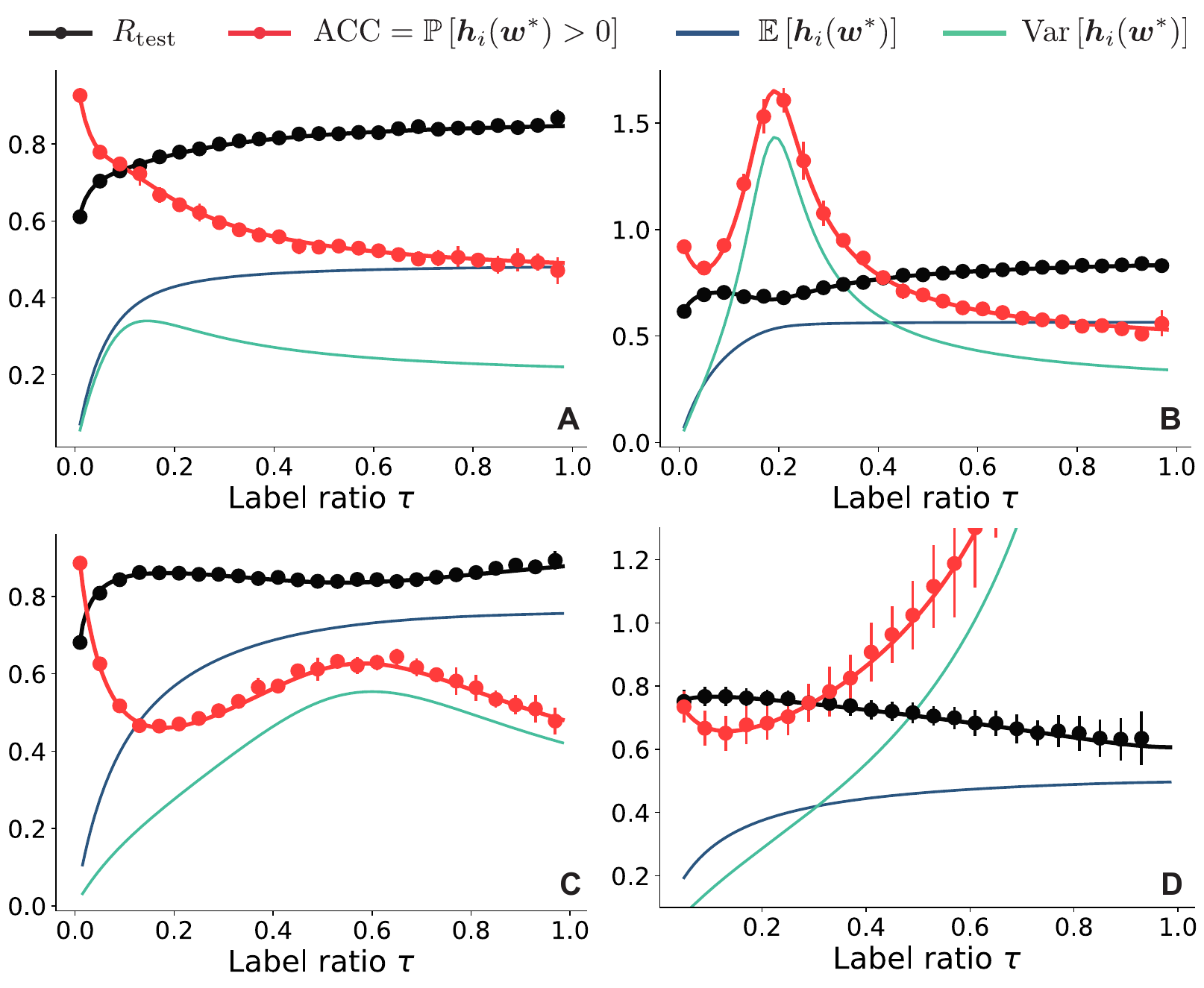}
    \caption{Four typical generalization curves in CSBM model. The solid lines represent theoretical results of test risk (black) and accuracy ({\textcolor{red}{red}}) computed via \eqref{eqn: Risk r>=0}.  We also plot the {\textcolor{myblue}{mean} and \textcolor{mygreen}{variance}} of test output $\vh_i(\vw^*)$ where $i\in V_{\text{test}}$. This illustrates how the tradeoff of Mean-Variance leads to different double descent curves. Note we only display results for nodes with label $\vy_i=1$; the result for the $\vy_i=-1$ class simply has opposite mean and identical variance.
     \textbf{\textsc{A}}: monotonic $\mathrm{ACC}$ (increasing) and $R_{\text{test}}$ (decreasing) when  regularization $r$ is large; \textbf{\textsf{B}}: A typical double descent with small regularization $r$; 
     \textbf{\textsf{C}} slight double descent with relative model complexity $\alpha$ close to $1$;
     \textbf{\textsf{D}} (near-monotonically) decreasing $\mathrm{ACC}$ and increasing $R_{\text{test}}$ with large relative model complexity $\alpha=1/\gamma$. The parameters are chosen as \textbf{\textsf{A}}: $\mu=1,\lambda=2,\gamma=5,r=2$; \textbf{\textsf{B}}: $\mu=1,\lambda=2,\gamma=5,r=0.1$;
     \textbf{\textsf{C}}: $\mu=1,\lambda=2,\gamma=1.2,r=0.05$; 
     \textbf{\textsf{D}}: $\mu=5,\lambda=1,\gamma=0.1,r=0.005$.
    %
    %
    The solid circles and vertical bars represent the mean and standard deviation of risk and accuracy from experiment results.
    Each experimental data point is averaged over $10$ independent trials; the standard deviation is indicated by vertical bars.  We use $N=5000$ and $d=30$ for \textbf{\textsf{A}}, \textbf{\textsf{B}} and \textbf{\textsf{C}}, and $N=500$ and $d=20$ for \textbf{\textsf{D}}. In all case we use the symmetric binary adjacency matrix ensemble $\mathcal{A}^{\text{bs}}$. } 
    \label{fig_acc}
\end{figure}


\subsection*{Heterophily, homophily, and positive and negative self-loops}

GCNs often perform worse on heterophilic than on homophilic graphs. An active line of research tries to understand and mitigate this phenomenon with special architectures and training strategies \cite{li2022finding,luan2022revisiting,chien2021adaptive}. We now show that it can be understood through the lens of self-loops.

\begin{figure*}[h!]
    \includegraphics[width=\linewidth]{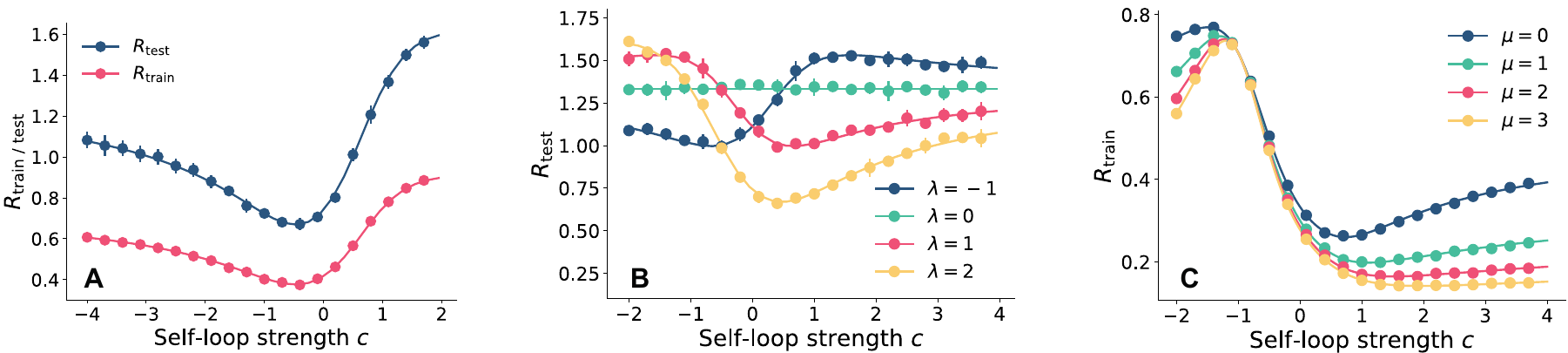}
    \caption{Train and test risks on CSBM for different intensities of self loops. \textbf{\textsf{A}}: train and test risk for $\tau=0.8$ and $\lambda=-1$ (heterophilic). \textbf{\textsf{B}}: test risks for $\gamma=0.8$, $\tau=0.8$, $\mu=0$ under different $\lambda$. \textbf{\textsf{C}}: training risk for different $\mu$ when $\tau=\lambda=1$. Each data point is averaged over $10$ independent trials with $N=5000$, $r=0$, and $d=30$. We use the non-symmetric binary adjacency matrix ensemble $\mathcal{A}^{\text{bn}}$. The solid lines are the theoretical results predicted by the replica method. In \textbf{\textsf{B}} we see that the optimal generalization performance requires adapting the self-loop intensity $c$ to the degree of homophily.} 
    \label{fig_selfloop}
\end{figure*}
    
Strong GCNs ubiquitously employ self-loops of the form $\mP(\mA) = \mA + \mI_N$ on homophilic graphs \cite{kipf2017semisupervised,chien2021adaptive,velivckovic2017graph,wu2019simplifying,gasteiger2018predict}.\footnote{One way to characterize link semantics in graphs is by notions of homophily and heterophily. In a friendship graph links signify similarity: if Alice and Bob both know Jane it is reasonable to expect that Alice and Bob also know each other. In a protein interaction graph, if proteins \texttt{A} and \texttt{B} interact, a small mutation \texttt{A}' of \texttt{A} will likely still interact with \texttt{B} but not with \texttt{A}. Thus ``interaction'' links signify partition. Most graphs are somewhere in between the homophilic and heterophilic extremes.} Self-loops, however, deteriorate performance on heterophilic networks. CSBM is well suited to study this phenomenon since $\lambda$ allows us to transition between homophilic and heterophilic graphs. 

We allow the self-loop strength $c$ to vary continuously so that the effective adjacency matrix becomes $\mA + c\mI_N$. Importantly, we also allow $c$ to be negative (see SI Appendix \ref{APP: Selfloop}). In Fig.~\ref{fig_selfloop} we plot the test risk as a function of $c$ for both positive and negative $c$. We find that a negative self-loop ($c < 0$) results in much better performance on heterophilic data ($\lambda < 0$). We sketch a signal-processing interpretation of this phenomenon in SI Appendix \ref{APP: selfloops and sp}.


\subsection*{Negative self-loops in state-of-the-art GCNs}

\begin{figure}[h!]
    \includegraphics[width=\linewidth]{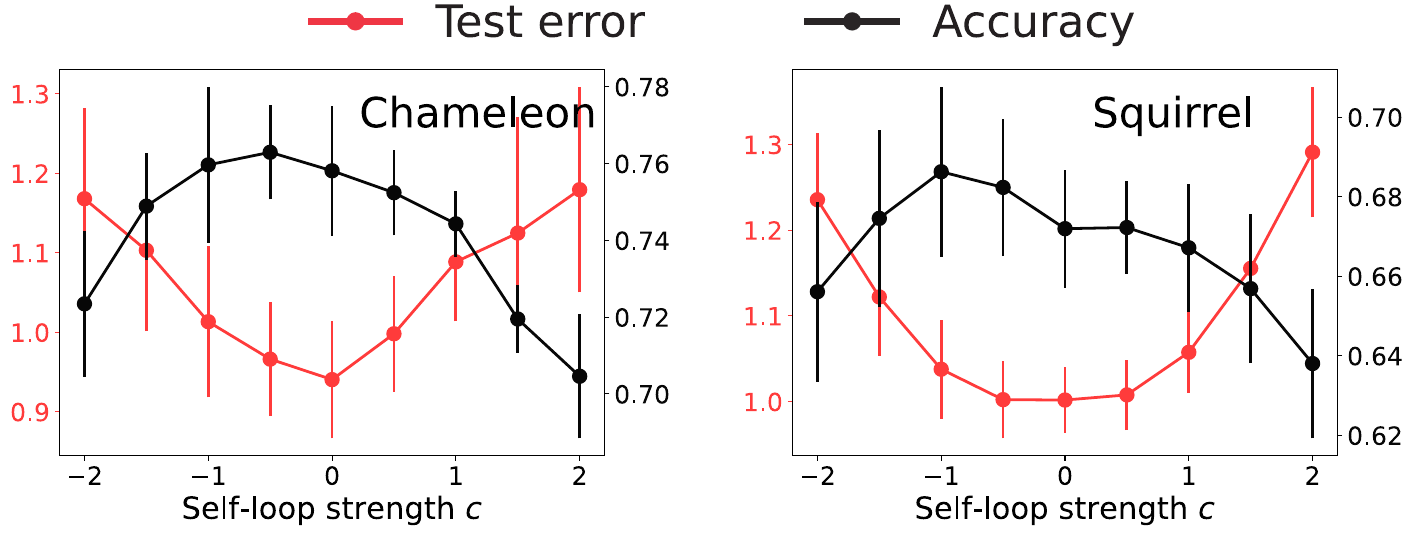}
    \caption{Test accuracy (\textcolor{black}{\textbf{black}}) and test error (\textcolor{red}{\textbf{red}}) in node classification with GCNs on real heterophilic graphs with different self-loop intensities. We implement a two-layer ReLU GCN with $128$ hidden neurons and an additional self-loop with strength $c$. Each setting is averaged over different training--test splits taken from \cite{pei2020geom} (60\% training, 20\% validation, 20\% test). The relatively large standard deviation (vertical bars) is  mainly due to the randomness of the splits. The randomness from model initialization and training is comparatively small. The optimal test accuracy for these two datasets is obtained for self-loop intensity $-0.5<c^*<-1$.} 
    \label{fig_selfloop_real}
\end{figure}

It is remarkable that this finding generalizes to complex state-of-the-art graph neural networks and datasets. We experiment with two common heterophilic benchmarks, \texttt{Chameleon} and \texttt{Squirrel}, first with a two-layer ReLU GCN. The default GCN (for example in \texttt{pytorch-geometric}) contains self-loops of the form $\mA + \mI$; we immediately observe in Fig.~\ref{fig_selfloop_real} that removing them improves performance on both datasets. We then make the intensity of the self-loop adjustable as a hyper-parameter and find that a negative self-loop with $c$ between $-1.0$ and $-0.5$ results in the highest accuracy on both datasets. It is notable that the best performance in the two-layer ReLU GCN with $c=-0.5$ (76.29\%) is already close to state-of-the-art results by the Feature Selection Graph Neural Network (FSGNN) \cite{maurya2021improving} (78.27\%). FSGNN uses a graph filter bank $\mathcal{B}=\{\mA^k,(\mA+\mI)^k\}$ with careful normalization. Taking a cue from the above findings, we show that a simple addition of negative self-loop filters $(\mA-0.5\mI)^k$ to FSGNN yields the new state of the art (78.96\%); see also Table \ref{tab:fsgnn}.

\begin{table}[h!]
\def\arraystretch{1.2}
\centering
\begin{tabular}{@{}lcccc@{}}
\toprule
          & \textbf{GCN} ($c=0$)  & \textbf{GCN} ($c^*$) & \textbf{FSGNN} & \textbf{FSGNN} ($c^*$) \\ \hline
\textbf{Chameleon} & 75.81±1.69 & 76.29±1.22 & 78.27±1.28& 78.96±1.05         \\
\textbf{Squirrel} & 67.19±1.48 & 68.62±2.13 & 74.10±1.89& 74.34±1.21   \\
\bottomrule
\end{tabular}
\caption{Comparison of test accuracy when negative self-loop is absent (first and third column) or present (second and fourth column). The datasets and splits are the same as Fig.~\ref{fig_selfloop_real}.}
\label{tab:fsgnn}
\end{table}

\section{Discussion}\label{sec: discussion}

Before delving into the details of the analytical methods in Section \ref{sec: main techniques} and conceptual connections between GNNs and spin glasses, we discuss the various interpretations of our results in the context of related work.

\subsection*{Related work on theory of GNNs}

Most theoretical work on GNNs addresses their expressivity  \cite{sato2020survey,geerts2022expressiveness}. A key result is that the common message-passing formulation is limited by the power of the Weisfeiler--Lehman graph isomorphism test \cite{xu2018how}. This is of great relevance for computational chemistry where one must discriminate between the different molecular structures \cite{gilmer2017neural}, but it does not explain how the interaction between the graph structure and the node features leads to generalization. Indeed, simple architectures like graph convolution networks (GCNs) are far from being universal approximators but they often achieve excellent performance on real problems with real data.

Existing studies of generalization in GNNs leverage complexity measures such as the Vapnik--Chervonenkis dimension \cite{vapnik1971uniform,vapnik1999nature,scarselli2018vapnik} or the Rademacher complexity \cite{garg2020generalization}. While the resulting bounds sometimes predict coarse qualitative behavior, a precise characterization of relevant phenomena remains elusive. Even the more refined techniques like PAC-Bayes perform only marginally better \cite{liao2021a}. It is striking that only in rare cases do these bounds explicitly incorporate the interaction between the graph and the features \cite{esser2021learning}. Our results show that understanding this interaction is crucial to understanding learning on graphs. 

Indeed, recall that a standard practice in the design of GNNs is to build (generalized) filters from the adjacency matrix or the graph Laplacian and then use these filters to process data. But if the underlying graph is an Erdős--Rényi random graph, the induced filters will be of little use in learning. The key is thus to understand how much useful information the graph provides about the labels (and vice-versa), and in what way that information is complementary to that contained in the features.

\subsection*{A statistical mechanics approach: precise analysis of simple models}

\idd{
An alternative to the typically vacuous\footnote{We quote the authors of the PAC-Bayesian analysis of generalization in GNNs \cite{liao2021a}: ``[...] we are far from being able to explain the practical behaviors of GNNs.''} complexity-based risk bounds for graph neural networks \cite{garg2020generalization,liao2021a,esser2021learning} is to adopt a statistical mechanics perspective on learning; this is what we do here. Indeed, one key aspect of learning algorithms that is not easily captured by complexity measures of statistical learning theory is the emergence of qualitatively distinct phases of learning as one varies certain key ``order parameters''. Such phase diagrams emerge naturally when one views machine learning models in terms of statistical mechanics of learning \cite{yang2021taxonomizing,martin2017rethinking}. 

Martin and Mahoney \cite{martin2017rethinking} demonstrate this elegantly by formulating what they call a \emph{very simple deep learning model}, and showing that it displays distinct learning phases reminiscent of many realistic, complex models, despite abstracting away all but the essential ``load-like'' and ``temperature-like'' parameters. They argue that such parameters can be identified in machine learning models across the board.

The statistical mechanics paradigm requires one to commit to a specific model and do different calculations for different models \cite{watkin1993statistical}, but it results in sharp characterizations of relevant phases of learning. 

Important results within this paradigm, both rigorous and heuristic, were derived over the last decade for regularized least-squares \cite{oymak2013squared, thrampoulidis2018precise,boyd2011distributed}, random-feature regression \cite{belkin2020two,liao2020random, hu2022universality, mei2022generalization}, and noisy Gaussian mixture and spiked covariance models \cite{el2018detection, macris2020all, mignacco2020role}, using a variety of analytical techniques from statistical physics, high-dimensional probability, and random matrix theory. Not all of these works \emph{explicitly} declare adherence to the statistical mechanics tradition. It nonetheless seems appropriate to categorize them thus since they provide precise analyses of learning in specific models in terms of a few order parameters.}

Even though these papers study comparatively simple models, many key results only appeared in the last couple of years, motivated by the proliferation of over-parameterized models and advances in analytical techniques. \idd{One should make sure to work in the correct scaling of the various parameters \cite{liao2020random}; while this may complicate the analysis it leads to results which match the behavior of realistic machine learning systems.} We extend these recent results by allowing the information to propagate on a graph; this gives rise to interesting new phenomena of some relevance for the practitioners. In order to obtain precise results we similarly study simple graph networks, but we also show that the salient predictions closely match the behavior of state-of-the-art networks on real datasets. We precisely traced the connection between generalization, the interaction type (homophilic or heterophilic) and the parameters of the GCN architecture and the dataset for a \textit{specific} graph model. Experiments show that the learned lessons apply to a broad class of GNNs and can be used constructively to improve the performance of state-of-the-art graph neural networks on heterophilic data.

\idd{Finally, let us mention that phenomenological characterizations of phase diagrams of risk are not the only way to apply tools from statistical mechanics and more broadly physics to machine learning and neural networks. These tools may help address a rather different set of ``design'' questions, as reviewed by Bahri et al. \cite{bahri2020statistical}.}

\subsection*{Negative self-loops in other graph learning models}

Recent theoretical work \cite{wei2022understanding,baranwal2023optimality} shows that optimal message passing in heterophilic datasets requires aggregating neighbor messages with a sign opposite from that of node-specific updates. Similarly, in earlier GCN architectures such as GraphSAGE \cite{hamilton2017inductive}, node and neighbor features are extracted using different trainable functions. This immediately allows the possibility of aggregating neighbors with an opposite sign in heterophilic settings. We show that self-loops with sign and strength depending on the degree of heterophily improve performance both in theory and in real state-of-the-art GCNs. The notion of self-loops in the context of GCNs usually indicates an explicit connection between a node and itself, $\mA \gets \mA + 
\mI$.

\subsection*{GCNs with a few labels outperform optimal unsupervised detection}

\begin{figure}[t!]
    \centering
    \includegraphics[width=0.8\linewidth]{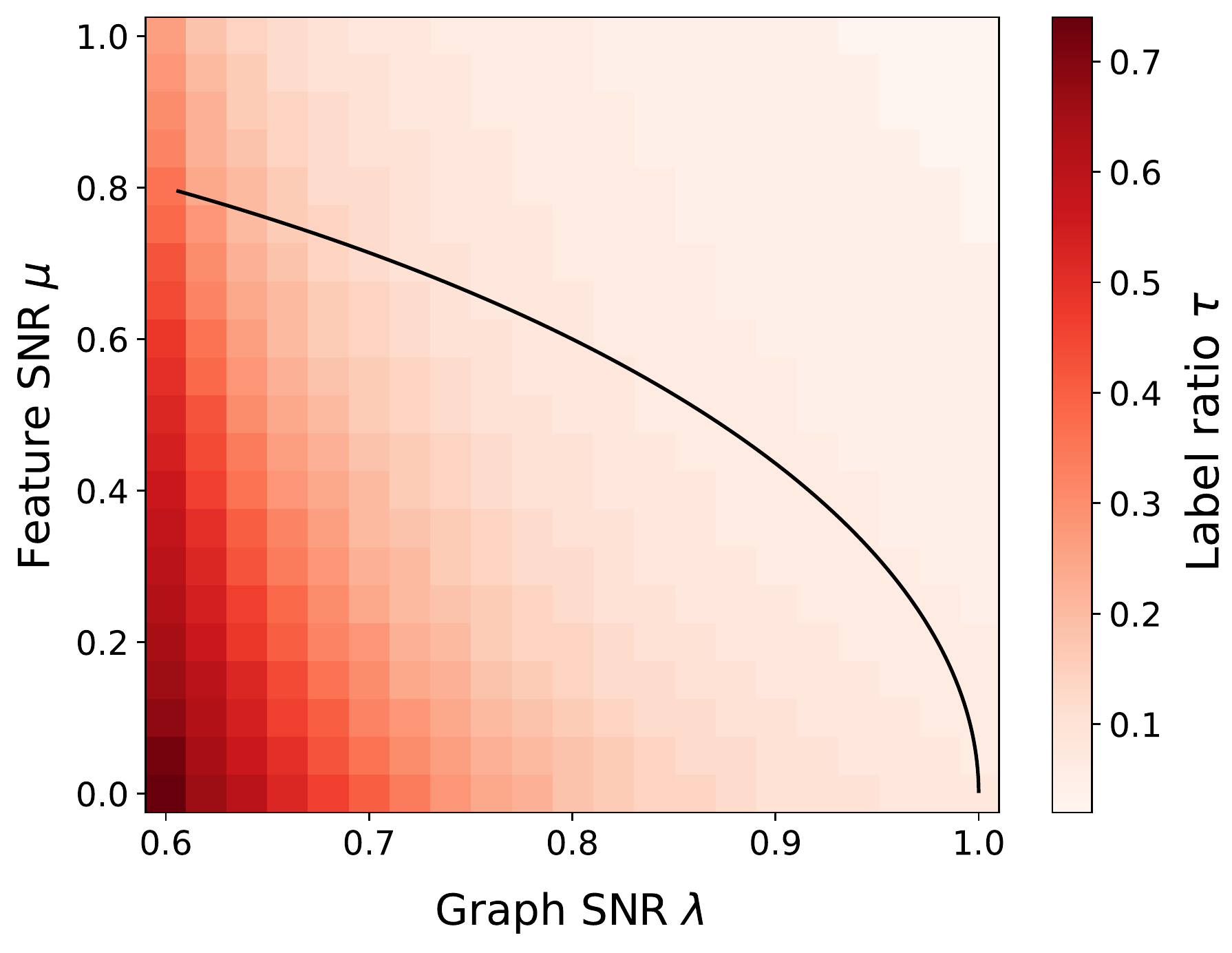}
    \caption{Training label ratio when a one-layer GCN matches the performance of unsupervised belief propagation at $\mu=\lambda=\gamma=1$. The black solid line denotes the information-theoretic detection threshold in the unsupervised setting where no label information is available ( i.e., when we use only $\mA$, $\mX$). If given a small number of labels, a simple, generally sub-optimal estimator matches the performance of the optimal unsupervised estimator.}
    \label{fig:super_vs_unsuper_heatmap}
\end{figure}

One interpretation of our results is that they quantify the value of labels in community detection, traditionally approached with unsupervised methods. These approaches are subject to fundamental information-theoretic detection limits which have drawn considerable attention over the last decade \cite{deshpande2017asymptotic,mossel2018proof,el2018detection}. The most challenging and most realistic high-dimensional setting is when the signal strength is comparable to that of the noise for both the graph and the features \cite{deshpande2018contextual, deshpande2017asymptotic,duranthon2023optimal}. The results of Deshpande et al. indicate that when $\mu^2 / \gamma +\lambda^2 < 1$, no unsupervised estimator can detect the latent structure $\vy$ from $\mA$ and $\mX$ \cite{deshpande2018contextual}. Our analysis shows that even a small fraction of revealed labels allow a simple GCN to break this \emph{unsupervised} barrier. 

In Fig.~\ref{fig:super_vs_unsuper_heatmap}, we compare the accuracy of a one-layer GCN with unsupervised belief propagation (BP) \cite{deshpande2018contextual}. We first run BP with $\mu=\lambda=\gamma=1$ and record the achieved accuracy. We then plot the smallest training label ratio $\tau$ for which the GCN achieves the same accuracy. We repeat this procedure for different feature SNRs $\mu$ and graph SNRs $\lambda$. The black solid line indicates the information-theoretic threshold for detecting the latent structure from $\mA$ and $\mX$. 

Earlier analyses of belief propagation in the SBM without features uncover a detectability phase transition \cite{zhang2014phase}. Our analysis shows that no such transition happens with GCNs. Indeed, our primary interest is in understanding GCNs, which are a general tool for a variety of problems, but unlike belief propagation, GCNs need not be near-optimal for community detection. For the optimal inference strategy, the phase transition may not be destroyed by revealing labels.


\section{Generalization in GCNs via statistical physics}\label{sec: main techniques}

The optimization problem \eqref{eqn: optim} has a unique minimizer as long as $r > 0$. Since it is a linear least-squares problem in $\vw$, we can write down a closed-form solution,
\begin{equation}\label{eqn: w star}
    \vw^*= (r\mI_F+ (\mP(\mA) \mX)^T \mI_{\text{train}} \mP(\mA) \mX)^{-1} (\mP(\mA)\mX)^T \mI_{\text{train} }\vy,
\end{equation}
where
\begin{equation}\label{eqn: Itrain}
\left(\mI_{\text{train}}\right)_{ij}=\begin{cases}
    1 & \text{if} \quad i=j\in V_{\text{train}}\\
    0 & \text{otherwise}.
\end{cases}
\end{equation}

Analyzing generalization is, in principle, as simple as substituting the closed-form expression \eqref{eqn: w star} into \eqref{eqn: risk} and \eqref{eqn: ACC} and calculating the requisite averages. The procedure is, however, complicated by the interaction between the graph $\mA$ and the features $\mX$ and the fact that $\mA$ is a random binary adjacency matrix. Further, for a symmetric $\mA$, $\mI_{\text{train}}\mP(\mA)$ is correlated with $\mI_{\text{test}}\mP(\mA)$ even in a shallow GCN (and certainly in a deep one).

\subsection*{The statistical physics program}

We interpret the (scaled) loss function as an energy, or a Hamiltonian, $H(\vw;\mA,\mX)= \tau N L_{\mA,\mX}(\vw)$. Corresponding to this Hamiltonian is the Gibbs measure over the weights $\vw$,
\begin{equation*}
\begin{aligned}
    \mathrm{d} \mathbb{P}_\beta(\vw;\mA,\mX)&=\frac{\exp(-\beta H(\vw;\mA,\mX)) \mathrm{d}\vw}{Z_{\beta}(\mA,\mX)} \\ \quad \text{where}\quad  Z_{\beta}(\mA,\mX)&=\int \mathrm{d} \vw \exp\left(-\beta H\left(\vw;\mA,\mX\right)\right),
\end{aligned}
\end{equation*}
$\beta$ is the \emph{inverse temperature} and $Z_\beta$ is the \emph{partition function}. At infinite temperature ($\beta \to 0$), the Gibbs measure is diffuse; as the temperature approaches zero $(\beta \to \infty)$, it converges to an atomic measure concentrated on the unique solution of \eqref{eqn: optim}, 
$
    \vw^*=\lim_{\beta\to\infty}   \int \vw \mathbb{P}_\beta(\vw;\mA,\mX) \mathrm{d} \vw.
$
In this latter case the partition function is similarly dominated by the minimum of the Hamiltonian. 
The expected loss can thus be computed from the \emph{free energy density} $f_\beta$,
\begin{equation*}
\begin{aligned}
\mathbb{E}_{\mA,\mX}[L_{\mA,\mX}(\vw^*)]=&-\frac{1}{\tau}\lim_{\beta\to \infty} f_\beta\\
\text{where} \quad f_\beta:=&-\lim_{N\to\infty}\frac{1}{N\beta}\mathbb{E}_{\mA,\mX}\ln Z_\beta(\mA,\mX).
\end{aligned}
\end{equation*}
Since the quenched average $\mathbb{E}\ln Z_{\beta}$ is usually intractable, we apply the replica method \cite{mezard1987spin} which allows us to take the expectation inside the logarithm and compute the annealed average,
\begin{equation*}
    \mathbb{E}_{\mA,\mX} \ln Z_\beta \left(\mA,\mX\right)  =\lim_{n \to 0 } \frac{\ln \mathbb{E}_{\mA,\mX}{Z_\beta^n} \left(\mA,\mX\right)}{n}.
\end{equation*}
The gist of the replica method is to compute $\mathbb{E}_{\mA,\mX}{Z_\beta^n}$ for integer $n$ and then ``pretend'' that $n$ is real and take the limit $n \to 0$. The computation for integer $n$ is facilitated by the fact that $Z_\beta^n$ normalizes the joint distribution of $n$ independent copies of $\vw$, $\set{\vw^a}_{a = 1}^n$. We obtain
\begin{equation}\label{eqn: z short}
\begin{aligned}
  &\mathbb{E}_{\mA,\mX}Z^{n}_{\beta}(\mA,\mX)\\
  &=\mathbb{E}_{\mA,\mX}\left(Z_{\beta}(\mA,\mX)\right)_{1}\times\cdots\times \left(Z_{\beta}(\mA,\mX)\right)_{n}\\
  &=\int\prod_{a=1}^{n}\mathrm{d}\vw^{a}\mathbb{E}_{\mA,\mX} \exp\left(\sum_{a=1}^{n} \left(-\beta\norm{\mI_{\text{train}} \mA\mX\vw^a-\mI_{\text{train}} \vy}_2^2\right) \right)\\
  &\hspace{3cm}\times \exp(-\beta \tau r \norm{\vw^a}_2^2).
\end{aligned}
\end{equation}

Instead of working with the product $\mA\mX$, replica allows us to express the free energy density as a stationary point of a function where the dependence on $\mA$ and $\mX$ is separated (see Appendix \ref{APP: Derivation Sketch} for details),
\begin{equation}\label{eqn: sketch saddle point}
\begin{aligned}
    f_\beta=&\frac{1}{\beta}\operatorname{extr}~ \lim_{n\to0}\lim_{N\to \infty} \frac{1}{nN} \left(\mathbb{E}_{\mA}\left[c\left(\mP(\mA)\right)\right]+\mathbb{E}_{\mX}\left[e(\mX)\right]\right)\\
    &+D(m,p,q,\widehat{m},\widehat{p},\widehat{q})\\
    =&\frac{1}{\beta}\underset{\substack{m,p,q \\ \widehat{m},\widehat{p},\widehat{q}}}{\operatorname{extr}}~ C(m,p,q)+E(\widehat{m},\widehat{p},\widehat{q})+D(m,p,q,\widehat{m},\widehat{p},\widehat{q}),
\end{aligned} 
\end{equation}
where we defined $C \bydef \frac{1}{nN}\mathbb{E}_{\mA}\left[c\left(\mP(\mA)\right)\right]$, $E \bydef \frac{1}{nN}\mathbb{E}_{\mX}\left[e(\mX)\right]$, which in the limit $N\to\infty$, $n\to 0$ only depend on the so-called order parameters $m,p,q$ and $\widehat{m},\widehat{p},\widehat{q}$. The separation thus allows us to study the influence of the distribution of $\mA$ in isolation; we provide the details in SI Appendix \ref{APP: Derivation Sketch}. The risks (called the observables in physics) can be obtained from $f_\beta$.  

\subsection*{Gaussian adjacency equivalence}

A challenge in computing the quantities in \eqref{eqn: z short} and \eqref{eqn: sketch saddle point} is to average over the binary adjacency matrix $\mA$. We argue that $f$ in \eqref{eqn: sketch saddle point} does not change if we instead average over the Gaussian ensemble with a correctly chosen mean and covariance. For a one-layer GCN ($\mP(\mA)=\mA$), we show that replacing $\mathbb{E}_{\mA^{\text{bs}}}c(\mP({\mA^{\text{bs}}}))$ by $\mathbb{E}_{\mA^{\text{gn}}} c\left(\mP({\mA^{\text{gn}}})\right)$ will not change $f$ in \eqref{eqn: sketch saddle point}with $\mA^{\text{gn}}$ being a spiked non-symmetric Gaussian random matrix,
\begin{equation}\label{eqn: A gn}
    \mA^{\text{gn}}=\frac{\lambda}{N}\vy \vy^T+\boldsymbol{\Xi}^{\text{gn}},
\end{equation}
with $\boldsymbol{\Xi}^{\text{gn}}$ having i.i.d. centered normal entries with variance $1/N$. This substitution is inspired by the universality results for the disorder of spin glasses \cite{talagrand2002gaussian,carmona2006universality,panchenko2013sherrington} and the universality of mutual information in CSBM \cite{deshpande2018contextual}. Deshpande et al. \cite{deshpande2018contextual} showed that the binary adjacency matrix in the stochastic block model can be replaced by  
\begin{equation}\label{eqn: A gs}
    \mA^{\text{gs}}=\frac{\lambda}{N}\vy\vy^T+\boldsymbol{\Xi}^{\text{gs}},
\end{equation}
where $\boldsymbol{\Xi}^{\text{gs}}\in\mathbb{R}^{N\times N}$ is a sample from the standard Gaussian orthogonal ensemble, without affecting the mutual information between $\vy$ (which they modeled as random) and ($\mA, \mX$) when $N \to \infty$ and $d \to \infty$.

Our claim refers to certain averages involving $\mA$; we record it as a conjecture since our derivations are based on the non-rigorous replica method. We first define four probability distributions:
\begin{itemize}
    \item $\mathcal{A}^{\mathrm{bs}}$: The distribution of adjacency matrices in the undirected CSBM (cf. \eqref{eqn: A bs}) scaled by $1/\sqrt{d}$, $\tfrac{1}{\sqrt{d}} \mA^{\mathrm{bs}} \sim \mathcal{A}^{\mathrm{bs}}$;
    \item $\mathcal{A}^{\mathrm{bn}}$: the distribution of adjacency matrices in the directed CSBM (cf. \eqref{eqn: A bn}), scaled by $1/\sqrt{d}$;
    \item $\mathcal{A}^{\mathrm{gs}}$: the distribution of spiked Gaussian orthogonal ensemble (cf. \eqref{eqn: A gs};
    \item $\mathcal{A}^{\mathrm{gn}}$: the distribution of spiked Gaussian random matrices (cf. \eqref{eqn: A gn}.
\end{itemize}

With these definitions in hand we can state
\begin{conjecture}[Equivalence of graph matrices] \label{conj: Equivalence}
Assume that $d$ scales with $N$ so that $1/d \to 0$ and $d/N\to 0$ when  $N\to\infty$. Let $\mP(\mA)$ be a polynomial in $\mA$ used to define the GCN function in \eqref{eq:lin-gnc}. It then holds that
$$
\begin{aligned}
    R_\mathrm{train}(\mathcal{A}^{\mathrm{b \bullet}})
    &= R_\mathrm{train}(\mathcal{A}^{\mathrm{g \bullet}}), \\
    R_\mathrm{test}(\mathcal{A}^{\mathrm{b \bullet}})
    &= R_\mathrm{test}(\mathcal{A}^{\mathrm{g \bullet}}), \\
    \mathrm{ACC}(\mathcal{A}^{\mathrm{b \bullet}}) 
    &= \mathrm{ACC}(\mathcal{A}^{\mathrm{g \bullet}}), \\
\end{aligned}
$$
with $\bullet \in \{ \mathrm{s}, \mathrm{n} \}$. When $\mP(\mA) = \mA$, the above quantities for symmetric and non-symmetric distributions also coincide.
\end{conjecture}

\begin{figure*}[ht!]
    \includegraphics[width=0.95\textwidth]{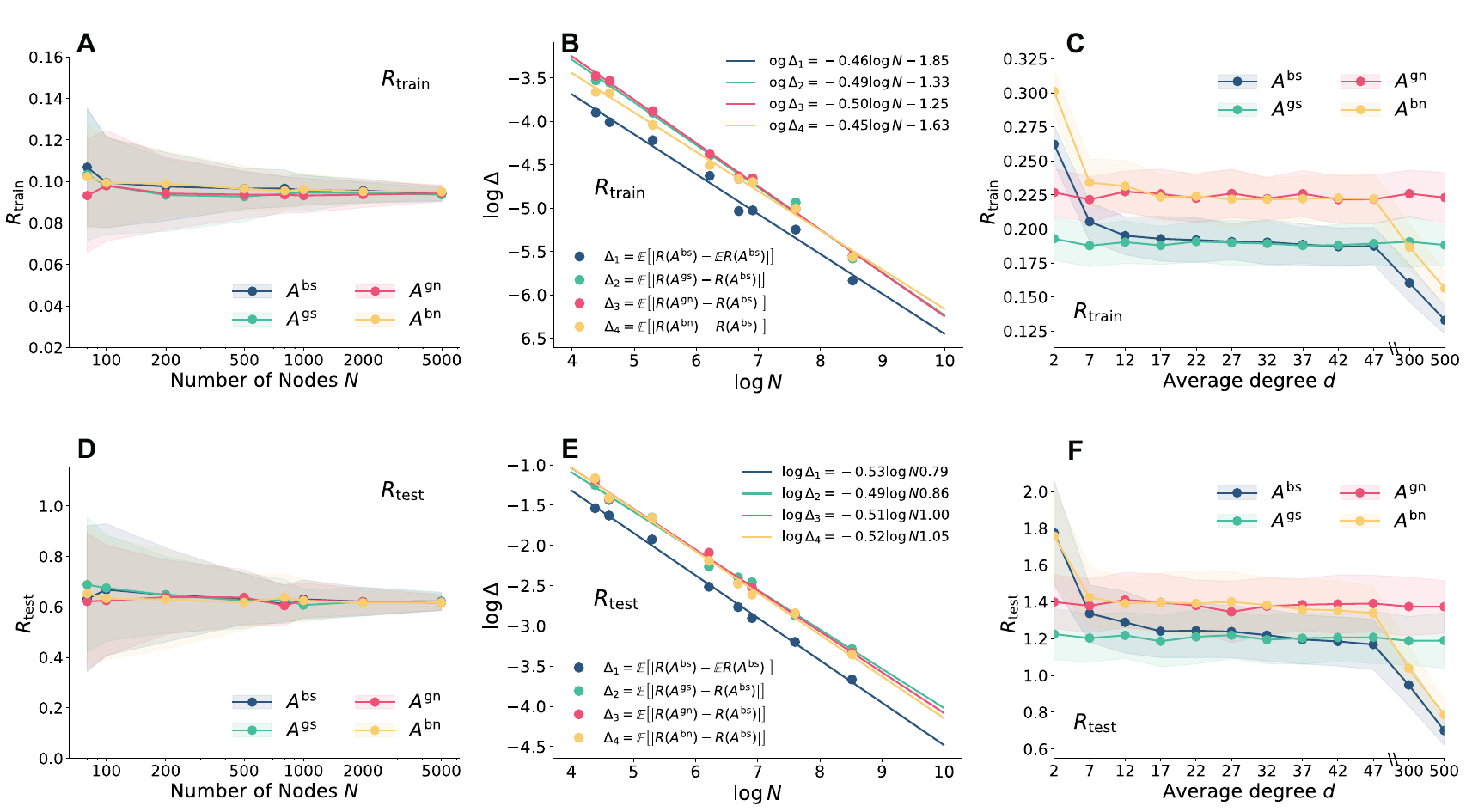}
    \caption{Numerical validation of Conjecture \ref{conj: Equivalence}. In \textbf{\textsf{A}} \& \textbf{\textsf{D}}: we show training and test risks with different numbers of nodes for $P(A)=A$. The parameters are set to $\gamma=\lambda=\mu=2,r=0.01,\tau=0.8$ and $d=\sqrt{N}/2$. In \textbf{\textsf{B}} \& \textbf{\textsf{E}}, we show the absolute difference of the risks between binary and Gaussian adjacency as a function of $N$, using the same data in \textbf{\textsf{A}} \& \textbf{\textsf{D}}. The solid lines correspond to a linear fit in the logarithmic scale, which shows that the error scales as $\vert\Delta \vert \sim N^{-0.5}$. In \textbf{\textsf{C}} \& \textbf{\textsf{F}} we show the training and test risks when  $\mP(\mA)=\mA^2$ under different average node degrees $d$. Other parameters are set to $\lambda=\mu=1,\gamma=2,N=2000, \tau=0.8$ and $r=0.01$. In these settings, the conjecture empirically holds up to scrutiny.} 
    \label{fig: conjecture}
\end{figure*}

In the case when $\mP(\mA) = \mA$ we justify Conjecture \ref{conj: Equivalence} by the replica method (see SI Appendix \ref{APP: Derivation Sketch}). In the general case we provide abundant numerical evidence in Fig.~\ref{fig: conjecture}. We first consider the case when $\mP(\mA)=\mA$. Fig.~\ref{fig: conjecture}\textbf{\textsf{A}} and Fig.~\ref{fig: conjecture}\textbf{\textsf{D}} show estimates of $R_\text{train}$ and $R_\text{test}$ averaged over $100$ independent runs. The standard deviation over independent runs is indicated by the shading. We see that the means converge and the variance shrinks as $N$ grows. 

We also show absolute differences between the averages of $R_\text{train}$ and $R_\text{test}$ in Fig.~\ref{fig: conjecture}\textbf{\textsf{B}} and  Fig.~\ref{fig: conjecture}\textbf{\textsf{E}}. We find that the values of $R_\text{train}$ and $R_\text{test}$ can be well fitted by a linear relationship in the logarithmic scale, suggesting that the absolute differences approach zero exponentially fast as $N \to \infty$. We next consider $\mP(\mA)=\mA^2$. In Fig.~\ref{fig: conjecture}\textbf{\textsf{C}} and Fig.~\ref{fig: conjecture}\textbf{\textsf{F}} we can see that for intermediate values of $d$, $R_\text{train}$ and $R_\text{test}$ corresponding to $\mA^{\text{bs}}$ and $\mA^{\text{bn}}$ are both close to that corresponding to $\mA^{\text{gs}}$ and $\mA^{\text{gn}}$. This is consistent with the results shown in Fig.~\ref{fig_v} and Fig.~\ref{fig_selfloop} where the theoretical results computed by the replica method and $\mA^{\text{gn}}$ perfectly match the numerical results with $\mA^{\text{bn}}$ (for $\mP(\mA)=\mA+c\mI_N$) and $\mA^{\text{bs}}$ (for $\mP(\mA)=\mA$), further validating the conjecture.

\subsection*{Solution to the saddle point equation}

We can now solve the saddle point equation \eqref{eqn: sketch saddle point} by averaging over $\mA^{\text{gn}}$. In the general case the solution is easy to obtain numerically. For an one-layer GCN with $\mP(\mA) = \mA$ we can compute a closed-form solution.
Denoting the critical point in \eqref{eqn: sketch saddle point} by $(m^*,p^*,q^*)$ we obtain
\begin{equation}\label{eqn: Risk r>=0}
\begin{aligned}
R_\text{train}&=\frac{(\lambda  m^*-1)^2+p^*}{(2 q^*+1)^2},
\\ R_\text{test}&= (\lambda  m^*-1)^2+p^*, \\ \mathrm{ACC}&=\frac{1}{2}\left(1+\operatorname{erf}\left(\frac{\lambda m^*}{\sqrt{2p^*}}\right)\right),
\end{aligned}
\end{equation}
where $\operatorname{erf}$ is the usual error function. While the general expressions are complicated (see SI Appendix \ref{APP: Derivation Sketch}), in the ridgeless limit $r \to 0$  we can compute simple closed-form expressions for train and test risks,
\begin{equation}\label{eqn: Risk r=0}
\begin{aligned}
R_\text{train}=& \frac{(\gamma+\mu)(\gamma\tau-1)}{\gamma\tau\left(\gamma+\lambda^{2}(\mu+1)+\mu\right)},\\
R_\text{test}=&\frac{\gamma\tau(\gamma+\mu)}{(\gamma\tau-1)\left(\gamma+\lambda^{2}(\mu+1)+\mu\right)},
\end{aligned}
\end{equation}
assuming that $\tau\gamma>1$.

\subsection*{A rigorous solution}

We note that for a one-layer GCN risks can be computed rigorously using random matrix theory provided that Conjecture \ref{conj: Equivalence} holds and we begin with a Gaussian ``adjacency matrix'' instead of the true binary SBM adjacency matrix. We outline this approach in SI Appendix \ref{APP: RMT}; in particular, for $r = 0$, the result of course coincides with that in \eqref{eqn: Risk r=0}.

\section{Conclusion}
\label{sec:conclusion}

We analyzed generalization in graph neural networks by making an analogy with a system of interacting particles: particles correspond to the data points and the interactions are specified by the adjacency relation and the learnable weights. The latter can be interpreted as defining the ``interaction physics'' of the problem. The best weights correspond to the most plausible interaction physics, coupled in turn with the network formation mechanism.

The setting that we analyzed is maybe the simplest combination of a graph convolution network and data distribution which exhibits interesting, realistic behavior. In order to theoretically capture a broader spectrum of complexity in graph learning we need to work on new ideas in random matrix theory and its neural network counterparts \cite{pennington2017nonlinear}. While very deep GCNs are known to suffer from oversmoothing, there exists an interesting intermediate-depth regime beyond a single layer \cite{keriven2022not}. Our techniques should apply simply by replacing $\mA$ by any polynomial $\mP(\mA)$ before solving the saddle point equation, but we will need a generalization of existing random matrix theory results for HCIZ integrals. 
Finally, it is likely that these generalized results could be made fully rigorous if  ``universality'' in Conjecture \ref{conj: Equivalence} could be established formally.

\acknow{We thank the anonymous reviewers for suggestions on how to improve presentation. Cheng Shi and Liming Pan would like to thank Zhenyu Liao (HUST) and Ming Li (ZJNU) for valuable discussions about RMT. Cheng Shi and Ivan Dokmani\'c were supported by the European Research Council (ERC) Starting Grant 852821---SWING. Liming Pan would like to acknowledge support from National Natural Science Foundation of China (NSFC) under Grant No.~62006122 and 42230406.
}

\showacknow{} 


\bibliography{pnas-sample}

\appendix

\newpage
\onecolumn
\section{Sketch of the derivation}\label{APP: Derivation Sketch}
\subsection{Replica method}\label{SEC: replica}

We now outline the replica-based derivations. We first consider one-layer GCN $\mP(\mA)=\mA$ to show the main idea. The training and test risks in this case is given by \eqref{eqn: risk} in the main text. We further extend the analysis when self-loops are included in Appendix \ref{APP: Selfloop}.   

We begin by defining the \emph{augmented partition function},
\begin{equation*}
Z_\beta(\mA,\mX)=\int\mathrm{d}\vw\exp\left(-\beta H(\vw) +t_0\beta O_\text{train}(\vw) +t_1 \beta O_\text{test}(\vw)\right),
\end{equation*}
where $\beta$ is the inverse temperature. The Hamiltonian in the above equation reads
\begin{equation}\label{eqn: Hami}
    \begin{aligned}
    H(\vw)&=\norm{\mI_{\text{train}}\mA\mX\vw-\mI_{\text{train}}\vy}_2^2+\tau r\norm{\vw}^2_2,
    \end{aligned}
\end{equation}
which is the loss \eqref{eqn: optim} scaled by $N\tau$. The ``observables'' $O_\text{train}$ and $O_\text{test}$ (the scaled training and test risks) are the quantities we are interested in:
\begin{equation*}
    O_\text{train}(\vw)=\norm{\mI_\text{train}\mA\mX\vw-\mI_\text{train}\vy}_2^2, \quad O_\text{test}(\vw)=\norm{\mI_\text{test} \mA\mX\vw-\mI_\text{test}\vy}_2^2.
\end{equation*}
When the inverse temperature $\beta$ is small, the Gibbs measure $Z_\beta^{-1}(A,X) \exp\left(-\beta H(\vw) \right) \mathrm{d}\vw$  is diffuse; when $\beta \to \infty$, the Gibbs measure converges to an atomic measure concentrated on the unique solution of \eqref{eqn: optim}. That is to say, for $t_0=t_1=0$, we can write
\begin{equation*}
\vw^*= \lim_{\beta \to \infty}\int \vw \, \mathbb{P}_\beta(\vw;\mA,\mX)\mathrm{d} \vw, \quad \text{where} \quad \mathbb{P}_\beta(\vw;\mA,\mX)=\frac{1}{Z_\beta(\mA,\mX)}\exp{\left(-\beta H\left(\vw\right)\right)}.
\end{equation*}
The idea is to compute the values of the observables in the large system limit at a finite temperature and then take the limit $\beta \to \infty$. To this end, we define the \emph{free energy density} $f_\beta$ corresponding to the augmented partition function,
\begin{equation}\label{eqn: free engery}
    f_\beta:= -\lim_{N\to\infty} \frac{1}{N\beta} \mathbb{E}_{\mA,\mX}\ln Z_\beta(\mA,\mX).
\end{equation}
The expected risks can be computed as
\begin{equation}\label{eqn: risk in f}
        R_{\text{train}} = -\frac{1}{\tau}\lim_{\beta \to \infty} \left. \frac{\partial f_\beta}{\partial t_0} \right \vert_{t_0=0,t_1=0}, \quad
    R_{\text{test}} = -\frac{1}{1-\tau}\lim_{\beta \to \infty} \left. \frac{\partial f_\beta}{\partial t_1} \right \vert_{t_0=0,t_1=0}.
\end{equation}
Although $\ln Z_{\beta}(\mA,\mB)/N$ concentrates for large $N$, a direct computation of the quenched average~\eqref{eqn: free engery} is intractable. We now use the \emph{replica trick} which takes the expectation inside logarithm, or replaces the quenched average by annealed average in physics jargon:
\begin{equation}\label{eqn: replica trick}
    \mathbb{E}_{\mA,\mX} \ln Z_\beta \left(\mA,\mX\right)  =\lim_{n \to 0 } \frac{\ln \mathbb{E}_{\mA,\mX}{Z_\beta^ n} \left(\mA,\mX\right)}{n}.
\end{equation}
The main idea of the replica method is to first compute $\mathbb{E}_{\mA,\mX}{Z_\beta^ n}$ for integer $n$ by interpreting $Z_\beta^n=(Z_\beta)_1(Z_\beta)_2\ldots(Z_\beta)_n$ as the product of $n$ partition functions for $n$ independent configurations $\set{\vw^a}_{a=1}^n$.  Then we obtain $\mathbb{E}_{\mA,\mX}\ln {Z_\beta}$ by taking the limit $n \to 0$ in \eqref{eqn: replica trick}, even though the formula for $\ln\mathbb{E}_{\mA,\mX}{Z_\beta}^n$ is valid only for integer $n$. The expectation of replicated partition function reads 
\begin{equation}\label{eqn: Z long}
\begin{aligned}
  \mathbb{E}_{\mA,\mX}Z^{n}_{\beta}(\mA,\mX)&=\mathbb{E}_{\mA,\mX}\left(Z_{\beta}(\mA,\mX)\right)_{1}\times\cdots\times \left(Z_{\beta}(\mA,\mX)\right)_{n}\\
  &=\int\prod_{a=1}^{n}\mathrm{d}\vw^{a}\mathbb{E}_{\mA,\mX} \exp\left(\sum_{a=1}^{n} \left(-\norm{\mI_\beta \mA\mX\vw^a-\mI_\beta \vy}_2^2\right) \right)\exp(-\beta \tau r \norm{\vw^a}_2^2),
\end{aligned}
\end{equation}
where we denote $\mI_\beta \bydef \sqrt{\beta-\beta t_0}\mI_{\text{train}}+\sqrt{-\beta t_1}\mI_{\text{test}}$. We first keep $\mX$ fixed and take the expectation over $\mA$. Directly computing the expectation over the binary graph matrix $\mA \sim \mathcal{A}^{\text{bs}}$ is non-trivial. To make progress, we average over $\mA \sim \mathcal{A}^{\text{gn}}$ (instead cf. \eqref{eqn: A gn}). In Appendix \ref{APP: C} we show that this Gaussian substitution does not change the free energy density, ultimately yielding the same risks and accuracies as detailed in Conjecture \ref{conj: Equivalence}.

Letting $\boldsymbol{\sigma}^a=\mX\vw^a$, the elements of  $\mA^{\text{gn}}\boldsymbol{\sigma}^a$ are now jointly Gaussian for any fixed $\boldsymbol{\sigma}^a$ and $C=\frac{1}{nN}\ln\mathbb{E}_{\mA}\exp\left(\sum_{a=1}^n\left(-\norm{\mI_\beta \mA\boldsymbol{\sigma}^a-\mI_\beta \vy}^2\right)\right)$ can be computed by multivariate Gaussian integration. It is not hard to see that $C$ depends only on a vector $\vm\in\R^{n}$ and a matrix $\mQ \in \R^{n\times n}$ defined as
\begin{equation}\label{eqn: order parameter}
    \vm_a=\vy^T \boldsymbol{\sigma}^a/N, \quad \text{and} \quad \mQ_{ab}=\left(\boldsymbol{\sigma}^a\right)^\intercal \boldsymbol{\sigma}^b/N.
\end{equation}
In statistical physics these quantities are called the \emph{order parameters}. We then define
\begin{equation}\label{eqn:replica-separated-C}
\begin{aligned}
    C(\vm,\mQ) &=\frac{1}{nN}\ln\mathbb{E}_{\mA} ~c(\mA) \\
    &=\frac{1}{nN} \ln\mathbb{E}_{\mA}\exp ~\left(-\sum_{a}\left\|\mI_{\beta} \mA \boldsymbol{\sigma}^{a}-\mI_{\beta} \vy\right\|_{2}^{2}\right).
\end{aligned}
\end{equation}
Using the Fourier representation of the Dirac delta function $\delta(t-t_0)=\frac{1}{2\pi}\int \mathrm{d}\omega \exp(i\omega(t-t_0))$, we have
\begin{equation}\label{eq:fourier-rep}
    \begin{aligned}
    \delta\left(N\mQ_{ab}-\left(\vw^{a}\right)^\intercal \mX^{\intercal}\mX\vw^{b}\right)&=\frac{1}{2\pi}\int d\widehat{\mQ}_{ab}\exp\left(i\widehat{\mQ}_{ab}\left(N\mQ_{ab}-\left(\vw^{a}\right)^{\intercal}\mX^{\intercal}\mX\vw^{b}\right)\right),\\\delta\left(N\vm_{a}-\vy^{\intercal}\mX\vw^{a}\right)&=\frac{1}{2\pi}\int d\widehat{\vm}_{a}\exp\left(i\widehat{\vm}_{a}\left(N\vm_{a}-\vy^{\intercal}\mX\vw^{a}\right)\right),
    \end{aligned}
\end{equation}
so that \eqref{eqn: Z long} becomes
\begin{equation}\label{eqn: Z C E}
\begin{aligned}
\mathbb{E}_{\mA,\mX}Z_\beta^{n}(\mA,\mX) & =\left(\frac{iN}{2\pi}\right)^{\frac{n^{2}+3 n}{2}}\int\mathrm{\prod_{a\leq b}\mathrm{d}\mQ_{ab}}\prod_{a\leq b}\mathrm{d}\widehat{\mQ}_{ab}\prod_{a}\mathrm{d}\vm_{a}\prod_{a}\mathrm{d}\widehat{\vm}_{a}\\&\quad\times\exp\left(nN C\left(\vm,\mQ\right)\right) \times \exp\left(nN E\left(\widehat{\vm},\widehat{\mQ}\right)\right)\\&\quad\times \exp\left(N\sum_{a\leq b}\widehat{\mQ}_{ab}\mQ_{ab}+N\sum_{a}\widehat{\vm}_{a}\vm_{a}\right),
\end{aligned}
\end{equation}
where 
\begin{equation}\label{eqn: replica sperated E}
\begin{aligned}
    E(\widehat{\vm},\widehat{\mQ}) 
    &= \frac{1}{nN} \ln \mathbb{E}_{\mX}~e(\mX)\\
    &=\frac{1}{nN}\ln \mathbb{E}_{\mX}\int\prod_{a=1}^{n}\mathrm{d}\vw_{a}\exp\bigg(-\sum_{a\leq b}\widehat{\mQ}_{ab}\left(\vw^{a}\right)^{T}\mX^{\intercal}\mX\vw^{b}-\sum_{a}\widehat{\vm}_{a}\vy^{\intercal}\mX\vw^{a}- \tau r\beta \sum_a \norm{\vw_a}_2^2\bigg).
    \end{aligned}
\end{equation}
In \eqref{eqn: Z C E} and \eqref{eqn: replica sperated E}, we apply the change of variables $i\mQ_{ab}\to\mQ_{ab}$,  $i\vm_{a}\to\vm_{a}$.
Note that while \eqref{eqn:replica-separated-C} still depends on $\mX$, we do not average over it.  As $\mA$ and $\mX$ appear as a product in \eqref{eqn:  replica trick}, it is not straightforward to average over them simultaneously. We average over $\mA$ first to arrive at \eqref{eqn: Z C E}, and then find that \eqref{eqn:replica-separated-C}-\eqref{eqn: replica sperated E} are self-averaging, meaning that they concentrate around their expectation as $N \to \infty$. Ultimately, this allows us to isolate the randomness in $\mA$ and $\mX$. It also allows the possibility to adapt our framework to other graph filters by simply replacing $\mA$ by $\mP(\mA)$ in \eqref{eqn:replica-separated-C}. Since \eqref{eqn: optim} has a unique solution, we take the replica symmetry assumption where the order parameters are structured as
\begin{equation}\label{eqn: replica sys}
    \vm=m{\bf{1}}_n,\quad  \widehat{\vm}=\widehat{m}{\bf{1}}_n, \quad \mQ=q \mI_n+ p  {\bf{1}}_n{\bf{1}}_n^\intercal,  \quad \quad \widehat{\mQ}= \widehat{q} 
    \mI_n+ \widehat{p}{\bf{1}}_n{\bf{1}}_n^\intercal.
\end{equation}
In the limit $N\to\infty$, $n\to 0$, we are only interested in the leading order contributions to $C(\vm,\mQ)$ so we write (with a small abuse of notation),
\begin{equation}\label{eqn: C m p q}
\begin{aligned}
        C(m,p,q)=&\frac{ \tau \left(2 \beta(1-t_0) q+\lambda m\right)^{2}}{2q\left(2 \beta(1-t_0) q+1\right)} -\beta(1-t_0)\tau  -\frac{ \tau}{2}\frac{2\beta(1-t_0) p}{2\beta(1-t_0) q+1}-\frac{ \tau \lambda^2 m^2}{2q}\\
        &+\frac{ \left(1-\tau\right)  \left(-2t_1 q+\lambda m\right)^{2}}{2q\left(-2t_1+1\right)} +t_1   \left(1-\tau\right)  -\frac{\left(1-\tau\right)}{2}\frac{-2t_1 p}{-2t_1 q+1} \\
    &-\frac{\left(1-\tau\right) \lambda^2 m^2}{2q}+ o(1)+\beta o_\beta(1).\footnotemark
\end{aligned}
\end{equation}
\footnotetext{We use the asymptotic notation $o(1)$ for deterministic and random quantities which vanish in the limit $n \to 0, N \to \infty$ in a suitable sense. We similarly use $o_\beta(1)$ in the limit  $\beta \to \infty$. The order parameters $m,p,q$ in \eqref{eqn: C m p q} as well as $\hat{m},\hat{p},\hat{q}$ in \eqref{eqn: App E} scale in linear order of $\beta$ or a polynomial of $\beta$.}
Similarly, we have
\begin{equation}\label{eqn: E}
\begin{aligned}
    E(\widehat{m},\widehat{p},\widehat{q})& = \frac{1}{2(2\widehat{q}+\widehat{p}) \gamma} \left(\widehat{p}\left(1- \widehat{r}T\right)+\gamma \widehat{m}^2 \left(1-\frac{T\widehat{r}+\gamma-1}{T^2 \mu \widehat{r} +(\gamma -1) T \mu +\gamma}\right) \right)+o(1)+\beta o_\beta(1),
\end{aligned}
\end{equation}
where
\begin{equation}
    T=\frac{1}{2\widehat{r}}\left(1-\widehat{r}-\gamma+\sqrt{1+\widehat{r}^{2}+2\widehat{r}+2\gamma \widehat{r}-2\gamma+\gamma^{2}}\right) \text{\quad and \quad} \widehat{r}=\frac{2\tau r}{2\widehat{q}+\widehat{p}}.
\end{equation}
We give the details of the derivation of $C$ in Appendix \ref{APP: C} and of $E$ in Appendix \ref{APP: E}. 

When $r\to 0$, we have $T\to \frac{1}{1-\gamma}$, and 
\begin{equation}\label{eqn: App E}
    E(\widehat{m},\widehat{p},\widehat{q})= \frac{1}{2}\left(\frac{ \widehat{m}^{2} }{2\widehat{q}+\widehat{p}}\right)\left(\frac{\mu+1}{\gamma+\mu}\right)-\frac{1}{2\gamma}\frac{\widehat{p}}{\left(2\widehat{q}+\widehat{p}\right)}+o(1).
\end{equation}
We can now compute \eqref{eqn: Z C E} by integrating only on $6$ parameters: $m$, $\widehat{m}$, $q$, $\widehat{q}$, $p$, and $\widehat{p}$. For $N\to\infty$, the integral can be computed via the saddle point method:
\begin{equation}\label{eqn: f beta}
    f_\beta=\frac{1}{\beta}\underset{\substack{m,\widehat{m},\\q,\widehat{q},p,\widehat{p}}}{\mathrm{extr}} \lim_{ n \to0}\lim_{N\to\infty}  C(m,p,q)+E(\widehat{m},\widehat{p},\widehat{q}) + \left(q+p\right)\left(\widehat{q}+\widehat{p}\right)-\frac{1}{2}p\widehat{p}+m\widehat{m}.
\end{equation}
The stationary point satisfies
\begin{equation}\label{eqn: APP sd1}
\frac{\partial f_{\beta}}{\partial m}=\frac{\partial f_{\beta}}{\partial\widehat{m}}=\frac{\partial f_{\beta}}{\partial p}=\frac{\partial f_{\beta}}{\partial\widehat{p}}=\frac{\partial f_{\beta}}{\partial q}=\frac{\partial f_{\beta}}{\partial\widehat{q}}=0.
\end{equation}
When $\beta \to \infty$, the stationary point exists only if $m,\widehat{m},p,\widehat{p},q,\widehat{q}$ scale as
\[
p=\mathcal{O}\left(1\right),~q=\mathcal{O}\left(\frac{1}{\beta}\right),~\widehat{p}=\mathcal{O}\left(\beta^{2}\right),~2\widehat{q}+\widehat{p}=\mathcal{O}\left(\beta\right),~m=\mathcal{O}\left(1\right),~\widehat{m}=\mathcal{O}\left(\beta\right).
\]
We thus reparameterize them as
\[
p\to p,~\beta q\to q,~\frac{\widehat{p}}{\beta^{2}}\to\widehat{p},~\frac{1}{\beta}\left(2\widehat{q}+\widehat{p}\right)\to\widehat{q},~m\to m,~\frac{1}{\beta}\widehat{m}\to\widehat{m}.
\]
Ignoring the small terms which vanish when $n \to 0, N \to \infty, \beta \to \infty$, and denoting $f=\lim _{\beta\to \infty} f_\beta$, we get
\begin{equation}
    \begin{aligned}\label{eqn: results f}
        f =\underset{\substack{m,\widehat{m},\\q,\widehat{q},p,\widehat{p}}}{\mathrm{extr}}  \quad  & g(t_0-1,\tau)+g(t_1,1-\tau) +\frac{1}{2}\left(\widehat{q}p+q\widehat{p}+2m\widehat{m}\right)\\
        &+\frac{1}{2}\frac{\widehat{m}^{2}}{\widehat{q}}\left(1-\frac{T \widehat{r}+\gamma-1}{T^{2} \mu \widehat{r}+(\gamma-1) T \mu+\gamma}\right)-\frac{1}{2\gamma}\frac{\widehat{p}\left(1- \widehat{r} T \right)}{\widehat{q}},
    \end{aligned}
\end{equation}
where
\begin{equation*}
g(t,\tau)=t \tau -\frac{\tau t p}{2 t q+1}-\frac{ \tau \lambda^2 m^2}{2 q}+\frac{ \tau \left(2 t q+\lambda m\right)^{2}}{2q\left(2 t  q+1\right)},
\end{equation*} and
$\widehat{r}=\frac{2 \tau r}{ \widehat{q}}$.
We denote by $m^*$, $p^*$, $q^*$, $\widehat{m}^*$, $\widehat{p}^*$, $\widehat{q}^*$ the stationary point in \eqref{eqn: results f} and substitute into \eqref{eqn: risk in f} yields the risks in \ref{eqn: Risk r>=0}.

We analyze the connection between the stationary point and $\vw^*$. 
As $\vw^*$ is the unique solution in \eqref{eqn: optim}, and from the definition of order parameters \eqref{eqn: order parameter}, stationarity in \eqref{eqn: results f} implies that
\begin{equation}\label{eqn: p* m*}
    \begin{aligned}
    p^*&=\lim_{N\to\infty} \mathbb{E}_{\mA,\mX} \left[{\vw^*}^T \mX^T \mX \vw^*/N\right],\\
    m^*&=\lim_{N\to\infty} \mathbb{E}_{\mA,\mX} \left[\vy^T \mX\vw^*/N\right].
    \end{aligned}
\end{equation}
Let $\mA_{\text{train}}\in \mathbb{R}^{F\times N}$ be the selection of rows from $\mA$ corresponding to $i$-th row for all $i \in V_\text{train}$ and $\mA_{\text{test}}\in \mathbb{R}^{(N-F)\times N}$ be the selection of rows  corresponding to $i$-th row for all $i \in V_\text{test}$. The neural network output for the test nodes reads $\vh_{\text{test}} = \mA_{\text{test}}\boldsymbol{\sigma}^*$, where $\boldsymbol{\sigma}^* = \mX\vw^*$.  Since we work with a non-symmetric Gaussian random matrix $\mA \sim \mathcal{A}^{\text{gn}}$ as our graph matrix, $\mA_\text{test}$ is independent of $\mA_\text{train}$ and $\boldsymbol{\sigma}^*$ (Note $\boldsymbol{\sigma}^*=X\vw^*$ depends on $\mA_\text{train}$). Therefore, for any fixed $\mA_{\text{train}}$ and $\mX$ but random $\mA_{\text{test}}$, the network outputs for test nodes are jointly Gaussian,
\begin{equation*}
    \mA_{\text{test}}\boldsymbol{\sigma}^* \sim \mathcal{N}\left({\frac{\lambda \vy^\intercal \boldsymbol{\sigma}^*}{ N}\vy, \left(\boldsymbol{\sigma}^*\right)^\intercal     \boldsymbol{\sigma}^* \boldsymbol{I}_N}\right).
\end{equation*}
Combining this with the results from \eqref{eqn: p* m*}, we obtain the test accuracy as
\begin{equation}
    \mathrm{ACC}=\mathbb{P}(x>0 ), \quad \text{where} \quad  x\sim \mathcal{N}\left(\lambda m^*, p^*\right)>0.
\end{equation}

\subsection{Computation of $C(\cdot)$}\label{APP: C}
Recall that in \eqref{eqn:replica-separated-C} we define
\begin{equation}\label{eqn: G}
    G\left(\mathcal{A}\right):=\frac{1}{nN}\ln \mathbb{E}_{\mA \sim \mathcal{A}} ~ c(\mA).
\end{equation}
In this section, we begin by computing $G(\mathcal{A}^{\text{gn}})=C(m,p,q)$ in \eqref{eqn: C m p q} where $\mathcal{A}^{\text{gn}}$ denotes the distribution of non-symmetric Gaussian spiked matrices \eqref{eqn: A gn}.  We then show that the symmetry does not influence the value of \eqref{eqn: G}, i.e., $G(\mathcal{A}^{\text{gn}})/{\beta}=G(\mathcal{A}^{\text{gs}})/{\beta}$ when $N,d,\beta \to \infty$ and $d/N,n \to 0$. Finally, we show that the Gaussian substitution for the binary adjacency matrix does not influence the corresponding free energy density, which ultimately leads to the same risks and accuracies under different adjacency matrices (Conjecture \ref{conj: Equivalence}).


Let's first concatenate $\{\boldsymbol{\sigma}^a\}_{a=1}^n$ as 
\begin{equation*}
    \Tilde{\boldsymbol{\sigma}}=\left[(\boldsymbol{\sigma}^1)_1, \cdots, (\boldsymbol{\sigma}^a)_1, \cdots, (\boldsymbol{\sigma}^n)_1,  (\boldsymbol{\sigma}^1)_2, \cdots, (\boldsymbol{\sigma}^a)_2, \cdots, (\boldsymbol{\sigma}^n)_2, \cdots, (\boldsymbol{\sigma}^n)_N\right]^\intercal.
\end{equation*}
Then we can rewrite \eqref{eqn: G} in vector form
\begin{equation}\label{eqn: G 2}
    G(\mathcal{A})=\frac{1}{nN} \ln \mathbb{E}_{\mA}\exp \left(-\left\|\left(\left(\mI_{\beta} \mA\right) \otimes \bm{1}_n\right) \Tilde{ \boldsymbol{\sigma}} -(\mI_{\beta}y) \otimes\bm{1}_n\right\|_{2}^{2}\right),
\end{equation}
where $\otimes$ is the Kronecker product. By the central limit theorem, when $N\to\infty$, the vectors $(\mA \otimes \boldsymbol{1}_n) \Tilde{\boldsymbol{\sigma}}$ for $\mA \sim\mathcal{A}^{\text{gn}}$, $\mA \sim \mathcal{A}^{\text{bn}}$, $\mA \sim \mathcal{A}^{\text{gs}}$ and $\mA \sim \mathcal{A}^{\text{bs}}$ all converge in distribution to Gaussian random vectors. Letting $\boldsymbol{\mu}(\mathcal{A})$ and $\boldsymbol{\Sigma}(\mathcal{A})$ be the mean and the covariance of $\mA\ \sim \mathcal{A}$, we get
\begin{equation}\label{eqn: G int}
    \begin{aligned}
        G(\mathcal{A})= & \frac{1}{nN}\Bigg(\ln\frac{1}{\sqrt{\det\left(\mI_{n N}+2\mI_{n \beta}^{2}{\boldsymbol{\Sigma}}\left(\mathcal{A}\right)\right)}}-\frac{1}{2}\boldsymbol{\mu}^{\intercal}\left(\mathcal{A}\right){\boldsymbol{\Sigma}}^{-1}\left(\mathcal{A}\right)\mu\left(\mathcal{A}\right)-\left(y\otimes \bm{1}_n\right)^{\intercal}\mI_{ n \beta}^{2}\left(\vy\otimes \bm{1}_n\right)\\&+\Big(\frac{1}{2}\left(2\left(\vy\otimes \bm{1}_n\right)^{\intercal}\mI_{n\beta}^{2}+\boldsymbol{\mu}^{\intercal}\left(\mathcal{A}\right){\boldsymbol{\Sigma}}^{-1}\left(\mathcal{A}\right)\right)  \left({\boldsymbol{\Sigma}}^{-1}\left(\mathcal{A}\right)+2\mI_{ n \beta}^{2}\right)^{-1}\\&\times \left(2\mI_{ n \beta}^{2}\vy\otimes \bm{1}_n+{\boldsymbol{\Sigma}}^{-1}\left(\mathcal{A}\right)\boldsymbol{\mu}\left(\mathcal{A}\right)\right)\Big)\Bigg)+o(1)
    \end{aligned}
\end{equation}
where $\mI_{n\beta}=\mI_\beta\otimes \mI_n$. The vanishing lower-order term $o(1)$ comes from the tails in the central limit theorem and it is thus absent when $\mathcal{A}=\mathcal{A}^{\text{gn}}$. In this case we have
\begin{equation}\label{ean: mu sigma gn}
    {\boldsymbol{\mu}}(\mathcal{A}^{\text{gn}})=\lambda \vy\otimes \vm, \quad {\boldsymbol{\Sigma}}(\mathcal{A}^{\text{gn}})=\mI_{N}\otimes \mQ,
\end{equation}
with $\vm$ and $\mQ$ defined in \eqref{eqn: order parameter}. Leveraging the replica symmetric assumption \eqref{eqn: replica sys}, we compute the determinant term in \eqref{eqn: G int} as
\begin{equation}\label{eqn: det C}
\begin{aligned}
    \frac{1}{nN}\det\left(\mI_{nN}+2\mI_{n\beta}^{2}{\boldsymbol{\Sigma}}\left(\mathcal{A}^{\text{gn}}\right)\right) &= \tau  \ln\left(2\beta\left(1-t_{0}\right)q+1\right)+\tau  \frac{2\beta\left(1-t_{0}\right) n p}{2\beta\left(1-t_{0}\right)q+1}\\
    &\ \ +\left(1-\tau\right) \ln\left(2\left(-\beta t_{1}\right)q+1\right)+\left(1-\tau\right) \frac{2\left(-\beta t_{1}\right)p}{2\left(-\beta t_{1}\right)q+1}\\
    &=\tau  \frac{2\beta\left(1-t_{0}\right)p}{2\beta\left(1-t_{0}\right)q+1}+\left(1-\tau\right) \frac{2\left(-\beta t_{1}\right)p}{2\left(-\beta t_{1}\right)q+1}+o(1)\\
    &\ \ +\tau  \ln\left(2\beta\left(1-t_{0}\right)q+1\right)+\left(1-\tau\right) \ln\left(2\left(-\beta t_{1}\right)q+1\right).
\end{aligned}
\end{equation}
The $o(1)$ in the third line comes from the approximation $\frac{1}{n}\ln(1+n)= 1+o(1)$. The last two terms in \eqref{eqn: det C} do not increase with $\beta$ and can thus be neglected in the limit $\beta \to \infty$ when computing $G(\mathcal{A})/\beta$: they give rise to $\beta o_\beta(1)$ in \eqref{eqn: C m p q}.  The rest terms in \eqref{eqn: G int} can be computed as 
\begin{equation*}
\begin{aligned}
    \frac{1}{2nN}\boldsymbol{\mu}\left(\mathcal{A}^{\text{gn}}\right)^{\intercal}\boldsymbol{\Sigma}^{-1}\left(\mathcal{A}^{\text{gn}}\right)\boldsymbol{\mu}\left(\mathcal{A}^{\text{gn}}\right)&=\frac{\lambda^{2} m^{2}}{2q}+o(1),\\
    \frac{1}{nN}\left(\vy\otimes \bm{1}_n\right)^{\intercal}\mI_{ n \beta}^{2}\left(\vy\otimes \bm{1}_n\right) &=\tau\beta\left(1-t_{0}\right)+\left(1-\tau\right)\left(-\beta t_{1}\right),\\
    \frac{1}{nN}\times\text{last two lines in \eqref{eqn: G int}}&=\tau\frac{\left(2\beta\left(1-t_{0}\right)q+\lambda m\right)^{2}}{2q\left(1+2\beta\left(1-t_{0}\right)q\right)}+\left(1-\tau\right)\frac{\left(2\left(-\beta t_{1}\right)q+\lambda m\right)^{2}}{2q\left(1+2\left(-\beta t_{1}\right)q\right)}+o(1).
\end{aligned}
\end{equation*}
Collecting everything we get $G(\mathcal{A}^{\text{gn}})$ in \eqref{eqn: C m p q}. Note that $G(\mathcal{A}^{\text{gn}})/\beta=\mathcal{O}(1)$, and we are going to show that $G(\mathcal{A}^{\text{gn}})/\beta-G(\mathcal{A}^{\text{gs}})/\beta=o(1)$.

For $\mathcal{A}^{\text{gs}},\mathcal{A}^{\text{bn}}$ and $\mathcal{A}^{\text{bs}}$, we find the means and covariances of $(\mA \otimes \bm{1}_n)\Tilde{\boldsymbol{\sigma}}$ as
\begin{equation}\label{eqn: mus and sigmas}
\begin{aligned}
\boldsymbol{\mu}(\mathcal{A}^{\text{gs}})&=\boldsymbol{\mu}(\mathcal{A}^{\text{gn}}),\quad &\boldsymbol{\Sigma}(\mathcal{A}^{\text{gs}})&= \boldsymbol{\Sigma}(\mathcal{A}^{\text{gn}})+\frac{1}{N}\Tilde{{\boldsymbol{\sigma}}}^{\intercal}\Tilde{{\boldsymbol{\sigma}}}, \\
\boldsymbol{\mu}(\mathcal{A}^{\text{bn}}) &=\boldsymbol{\mu}(\mathcal{A}^{\text{gn}}) +\sqrt{d} \boldsymbol{1}_N \otimes \vl,\quad &\boldsymbol{\Sigma}(\mathcal{A}^{\text{bn}})&=\boldsymbol{\Sigma}(\mathcal{A}^{\text{gn}})\left(1+\mathcal{O}\left(\frac{1}{\sqrt{d}}+\frac{d}{N}\right)\right),\\
\boldsymbol{\mu}(\mathcal{A}^{\text{bs}})&=\boldsymbol{\mu}(\mathcal{A}^{\text{gn}})+\sqrt{d} \boldsymbol{1}_N \otimes \vl ,\quad &\boldsymbol{\Sigma}(\mathcal{A}^{\text{bs}})&=\boldsymbol{\Sigma}(\mathcal{A}^{\text{gn}})\left(1+\mathcal{O}\left(\frac{1}{\sqrt{d}}+\frac{d}{N}\right)\right)+\frac{1+\mathcal{O}\left(\frac{1}{\sqrt{d}}+\frac{d}{N}\right)}{N}\Tilde{{\boldsymbol{\sigma}}}^{\intercal}\Tilde{{\boldsymbol{\sigma}}},
\end{aligned}
\end{equation}
where $\vl \in \mathbb{R}^{n}$ with entries $\vl_a=\frac{\bm{1}_N^\intercal \boldsymbol{\sigma}^a}{N}$ are order parameters analogous to $\vm$ in \eqref{eqn: order parameter}.  
Substituting \eqref{eqn: mus and sigmas} into \eqref{eqn: G int}, we see that the perturbation $\frac{1}{N}\Tilde{{\boldsymbol{\sigma}}}^{\intercal}\Tilde{{\boldsymbol{\sigma}}}$ in $\boldsymbol{\Sigma}(\mathcal{A}^{\text{gs}})$ leads to a $\mathcal{O}(\frac{1}{N})$ perturbation in $G(\mathcal{A}^{\text{gs}})$, while the perturbation of $\mathcal{O}\left(\frac{1}{\sqrt{d}}+\frac{d}{N}\right)$ in $\Sigma(\mathcal{A}^{\text{bn}})$ and $\Sigma(\mathcal{A}^{\text{bs}})$ leads to a $\mathcal{O}(\frac{1}{\sqrt{d}}+\frac{d}{N})$ perturbation in $G(\mathcal{A})$. 

In $\boldsymbol{\mu}(\mathcal{A}^{\text{bn}})$ and $\boldsymbol{\mu}(\mathcal{A}^{\text{gn}})$, there is a bias $\vl$. By the replica symmetric assumption \eqref{eqn: replica sys}, we have
$
    \vl = l \boldsymbol{1}_n.
$
It is easy to show that a critical point in the saddle point equation with $l$ and $\hat{l}$ exists only when $l=0$ for $\beta \to \infty$. Therefore, the term $\sqrt{d}\boldsymbol{1}_N\otimes\vl$ will not influence the value of free energy density when $\beta \to \infty$. It further implies that the elements of $\boldsymbol{\sigma}^*=\mX\vw^*$ are symmetrically distributed around zero.  This is analogous to the vanishing average magnetization for the Sherrington-Kirkpatrick model~\cite{mezard1987spin}. 

To summarize this section,  as long as $\frac{1}{N},\frac{1}{\sqrt{d}},  \frac{d}{N}\to 0$,  averaging over $\mathcal{A}^{\text{bs}}$, $\mathcal{A}^{\text{gn}}$, $\mathcal{A}^{\text{gs}}$ and $\mathcal{A}^{\text{gn}}$ are equivalent in \eqref{eqn: f beta} when $\beta \to \infty$ for a one-layer GCN with $\mP(\mA)=\mA$.\footnote{In general it does not hold that $G(\mathcal{A}^{\text{gn}})=G(\mathcal{A}^{\text{bn}})$ nor that $G(\mathcal{A}^{\text{gs}})=G(\mathcal{A}^{\text{bs}})$. The equivalence stems from the fact that $\vsigma^*=\mX\vw^*$ are symmetrically distributed, but for any fixed $\vsigma^*$ with $l=\boldsymbol{1}^\intercal_N \vsigma^* \neq0$, we have $G(\mathcal{A}^{\text{gn}})=G(\mathcal{A}^{\text{bn}})+l^2\mathcal{O}(1)$ and $G(\mathcal{A}^{\text{gs}})=G(\mathcal{A}^{\text{bs}})+l^2\mathcal{O}(1)$. }

\subsection{Computation of $E$}\label{APP: E}
We recall \eqref{eqn: replica sperated E} and denote $\bar{\mQ}=\widehat{\mQ}+\mathrm{diag}(\widehat{\mQ})$, then
\begin{equation}\label{eqn: APP exp E}
\begin{aligned}
    E(\widehat{\vm},\widehat{\mQ})=& \frac{1}{nN}\ln \mathbb{E}_{\mX} \sqrt{ \frac{\left(2\pi\right)^{n N}}{\det\left(\mX^{\intercal}\mX \otimes \bar{\mQ}+ 2 \tau r\beta \mI_{n F}\right)}}\\ &\times \exp\left(\frac{1}{2}\left(\vy^{\intercal}\mX \otimes\widehat{\mM}^{\intercal}\right)\left( \mX^{\intercal}\mX\otimes \bar{\mQ}+2 r \tau \beta \mI_{n F}\right)^{-1}\left(\mX^{\intercal}\vy\otimes\widehat{\mM}\right)\right).
\end{aligned}
\end{equation} 
We can compute the determinant term and the exponential term in \eqref{eqn: APP exp E} separately when $N\to\infty$. Denoting by $\lambda_f(\mX^\intercal \mX)$ the $f$-th largest eigenvalue of $\mX^\intercal \mX$, we have
\begin{equation}\label{eqn: E det}
\begin{aligned}
        &\frac{1}{nN }\ln \left(\sqrt{\frac{(2\pi)^{n N}}{\det \left(\mX^\intercal \mX \otimes  \widehat{\mQ}  +2 \tau r \beta \mI_{n F} \right)} }\right)\\
         = & -\frac{  \widehat{p} }{2\gamma \left(2\widehat{q}+\widehat{p}\right)} \left(1-\sum_f^F\frac{1}{F} \frac{2\tau r\beta/\left(2\widehat{q}+\widehat{p}\right)}{\lambda_{f}+ 2\tau r\beta/\left(2\widehat{q}+\widehat{p}\right)}\right)+\frac{1 }{2\gamma}\log\left(2\pi\right)-\frac{1}{F}\frac{n }{2\gamma}\sum_{f=1}^F\log\left(\left(2\widehat{q}+\widehat{p}\right) \lambda_f+2\tau r\beta\right)+o(1).
\end{aligned}
\end{equation}
For the same reasons as in \eqref{eqn: det C}, the two logarithmic terms in the last line of \eqref{eqn: E det} can be ignored since they do not grow with $\beta$. We denote $\widehat{r} \bydef 2\tau r\beta/\left(2\widehat{q}+\widehat{p}\right)$. The first term in the RHS of \eqref{eqn: E det} is then 
\begin{equation}\label{eqn: APP define T}
\begin{aligned}
T(\widehat{r},\gamma)
&= \frac{1}{\widehat{r}}\sum_f^F\frac{1}{F} \frac{\widehat{r}}{\lambda_{f} + \widehat{r}}\\
&= \frac{1}{2\widehat{r}}\left(1-\widehat{r}+\sqrt{1+\widehat{r}^{2}+2\widehat{r}+2\gamma \widehat{r}-2\gamma+\gamma^{2}}-\gamma\right)+o(1),
\end{aligned}
\end{equation}
which is obtained by integrating over the Marchenko--Pastur distribution. 

The term inside the exponential in \eqref{eqn: APP exp E} can be computed as
\begin{equation*}
\begin{aligned}
    & &&\frac{1}{2nN}\left( \vy^{\intercal} \mX \otimes \widehat{\vm}^{\intercal} \right)\left(  \mX^{\intercal} \mX \otimes \bar{\mQ}+2\tau r \beta \mI_{n F}\right)^{-1}\left(\mX^{\intercal}\vy \otimes  \widehat{\vm} \right)\\
    &=&&\frac{1}{2nN}\left( \vy^{\intercal} \mX \otimes \widehat{\vm}^{\intercal} \right)\left(  \mX^{\intercal} \mX \otimes \left(\bar{\mQ}-p{\mathbf{1}_n \mathbf{1}_n}^\intercal \right)+2\tau r \beta \mI_{n F}\right)^{-1}\left(\mX^{\intercal}  \vy \otimes \widehat{\vm} \right)+o(1)\\
    &=&&\frac{1}{2nN}\left(\widehat{\vm}^{\intercal} \left(\bar{\mQ}-\widehat{p}{\mathbf{1}_n\mathbf{1}_n}\right)^{-1} \widehat{\vm}\right)\left(\vy^{\intercal}\mX\left(\mX^{\intercal}\mX+\widehat{r}\mI_F\right)^{-1}\mX^{\intercal}\vy\right)+o(1)\\
    &=&&\frac{1}{2} \frac{\widehat{m}^2  }{2\widehat{q}+\widehat{p}}  \left(1-\frac{T\widehat{r}+\gamma-1}{T^2 \mu \widehat{r} +(\gamma -1) T \mu +\gamma}\right)+\title{o}(1),
    \end{aligned}
\end{equation*}
in which we first perturb $\bar{\mQ}$ by $p\boldsymbol{1}_n\boldsymbol{1}_n^\intercal$ in the second line and then compute $\left(\vy^{\intercal}\mX\left(\mX^{\intercal}\mX+\widehat{r}\mI_F\right)^{-1}\mX^{\intercal}\vy\right)$ by the Woodbury matrix identity.

\section{Self-loop computation}\label{APP: Selfloop}

When $\mP(\mA)=\mA+c\mI_N$, we still follow the replica pipeline from Section \ref{APP: Derivation Sketch} but with different (more complicated) $C$ and $E$ in \eqref{eqn: Z C E}.
\subsection{$C$ with self-loop} We replace $\mA$ by $\mA+c\mI_N$ in \eqref{eqn:replica-separated-C}. Now the expectation over $\mA$ depends not only on $\vm$ and $\mQ$, but also
\begin{equation}\label{eqn: new order para}
\begin{aligned}
    \left(\vm^0\right)_a&=&&\frac{1}{N}\vy^\intercal \mI_{\text{train}}\boldsymbol{\sigma}^a, & \left(\vm^1\right)_a&=&&\frac{1}{N}\vy^\intercal \mI_{\text{test}}\boldsymbol{\sigma}^a,\\
    \left(\mQ^0\right)_{ab}&=&&\frac{1}{N} \left(\boldsymbol{\sigma}^a\right)^\intercal \mI_{\text{train}}\boldsymbol{\sigma}^b, & \left(\mQ^1\right)_{ab}&=&&\frac{1}{N} \left(\boldsymbol{\sigma}^a\right)^\intercal \mI_{\text{test}}\boldsymbol{\sigma}^b.
\end{aligned}
\end{equation}

For $\mA \sim \mathcal{A}^{\text{gn}}$, since $((\mA+c\mI_N)\otimes \bm{1}_n)\boldsymbol{\Tilde{\sigma}}$ is a Gaussian with mean and covariance 
\begin{equation*}
    \boldsymbol{\mu}(\mathcal{A}^{\text{gn}})=\lambda \vy\otimes \vm+c\Tilde{\boldsymbol{\sigma}}, \quad \boldsymbol{\Sigma}(\mathcal{A}^{\text{gn}})=\mI_{N}\otimes \mQ,
\end{equation*}we can compute \eqref{eqn: G int} directly. Leveraging the replica symmetric assumption, i.e.,
\begin{equation*}
\begin{aligned}
        \vm^0&=m_0\boldsymbol{1}_n\quad &\vm^1&=m_1\boldsymbol{1}_n\\
        \mQ^0&=q_0\mI_n+ p_0 \boldsymbol{1}_n \boldsymbol{1}_n^\intercal \quad &\mQ^1&=q_1\mI_n+ p_1 \boldsymbol{1}_n \boldsymbol{1}_n^\intercal,\\
\end{aligned}
\end{equation*}
we have
\begin{equation*}
\begin{aligned}
   C(\vm^0,\vm^1,\mQ^0,\mQ^1) = & ~ \tau\frac{\left(2\beta_{0}q+\lambda m\right)^{2}}{q\left(1+2\beta_{0}q\right)}+\left(1-\tau\right)\frac{\left(2\beta_{1}q+\lambda m\right)^{2}}{q\left(1+2\beta_{1}q\right)}\\&+c^{2}\frac{\left(1+2\beta_{0}q\right)p_{0}q-\left(1+4\beta_{0}q\right)pq_{0}}{q^{2}\left(1+2\beta_{0}q\right)^{2}}+c^{2}\frac{\left(1+2\beta_{1}q\right)p_{1}q-\left(1+4\beta_{1}q\right)pq_{1}}{q^{2}\left(1+2\beta_{1}q\right)^{2}}\\&+2c\left(\frac{\lambda m+2\beta_{0}q}{q\left(1+2\beta_{0}q\right)}m_{0}+\frac{\lambda m+2\beta_{1}q}{q\left(1+2\beta_{1}q\right)}m_{1}\right)+o(1)+o_\beta(1).
\end{aligned}
\end{equation*}
where $\beta_0=\beta(1-t_0)$, $\beta_1=-\beta t_1$, $q=q_0+q_1$, $m=m_0+m_1$ and $p=p_0+p_1$.

\subsection{$E$ with self-loop}

After applying the Fourier representation method \eqref{eq:fourier-rep} to the order parameters \eqref{eqn: new order para}, we obtain dual variables $\widehat{\vm}^0,\widehat{\vm}^1,\widehat{\mQ}^0,\widehat{\mQ}^1$. We then get a new \eqref{eqn: replica sperated E} as
\begin{equation*}
\begin{aligned}
    E(\widehat{\vm}^0,\widehat{\vm}^1,\widehat{\mQ}^0,\widehat{\mQ}^1)=\frac{1}{nN}\ln \mathbb{E}_{\mX}\int\prod_{a=1}^{n}\mathrm{d}\vw_{a}\exp\left(-\sum_{a\leq b}(\widehat{\mQ}^0)_{ab}\left(\vw^{a}\right)^{\intercal}\mX^{\intercal}\mI_{\text{train}}\mX w^{b} -\sum_{a}(\widehat{\vm}^0)_{a}\vy^{\intercal}\mX\mI_{\text{train}} \vw^{a} \right.\\\left. -\sum_{a\leq b}(\widehat{\mQ}^1)_{ab}\left(\vw^{a}\right)^{\intercal}\mX^{\intercal}\mI_{\text{test}}\mX \vw^{b} -\sum_{a}(\widehat{\vm}^1)_{a}\vy^{\intercal}\mX\mI_{\text{test}}\vw^{a}-\tau r\beta \sum_a \norm{\vw^a}_2^2\right).
    \end{aligned}
\end{equation*}
Integrating over $\vw_a$ yields
\begin{equation}\label{APP: E selfloop}
\begin{aligned}
    E=&~ \frac{1}{nN}\ln \mathbb{E}_{\mX} \sqrt{\frac{(2\pi)^{n N}}{\mathrm{det}\left(\mX^{\intercal} \mI_{\text{train}}\mX \otimes \bar{\mQ}^{0}+\mX^{\intercal} \mI_{\text{test}}X\otimes\bar{\mQ}^{1}+2\tau r\beta \mI_{n F}\right)}}\\
    & ~ \times \exp\left(\frac{1}{2}\left(\vy^{\intercal} \otimes\left(\widehat{\vm}^{0}\right)^{\intercal}  \mI_{\text{train}}\mX+ \vy^{\intercal} \mI_{\text{test}}\mX \otimes \left(\widehat{\vm}^{1}\right)^{\intercal} \right) \right.\\
    & ~ \times\left(\mX^{\intercal} I_{\text{train}}\mX \otimes  \bar{\mQ}^{0}+\mX^{\intercal} \mI_{\text{test}}\mX \otimes \bar{\mQ}^{1} +2\tau r\beta \mI_{nF}\right)^{-1}  \left(\mX^{\intercal} \mI_{\text{train}}\vy \otimes \widehat{\vm}^{0}+\mX^{\intercal} \mI_{\text{test}}\vy\otimes\widehat{\vm}^{1}\right),
\end{aligned}
\end{equation}
where $\bar{\mQ}^0=\widehat{\mQ}^0+\mathrm{diag}(\widehat{\mQ}^0)$ and $\bar{\mQ}^1=\widehat{\mQ}^1+\mathrm{diag}(\widehat{\mQ}^1)$. By replica symmetry, we have
\begin{equation}
\begin{aligned}
        \widehat{\vm}^0&=\widehat{m}_0\boldsymbol{1}_n \quad &\widehat{\vm}^1&=\widehat{m}_1\boldsymbol{1}_n\\
        \bar{\mQ}_0&=\widehat{q}_0\mI_n+ \widehat{p}_0 \boldsymbol{1}_n \boldsymbol{1}^\intercal_n \quad &\bar{\mQ}_1&=\widehat{q}_1\mI+ \widehat{p}_1 \boldsymbol{1}_n \boldsymbol{1}_n^\intercal.\\
\end{aligned}
\end{equation}
Similarly as in Appendix \ref{APP: E}, we can compute the determinant term and the exponential term separately when $N\to\infty$. For the sake of simplicity, we only consider $r=0$ and $\mu=0$ in what follows. Denoting $\mY^0=\mX^{\intercal} \mI_{\text{train}}\mX$ and $\mY^1=\mX^{\intercal} \mI_{\text{test}}\mX$, we write the determinant term in \eqref{APP: E selfloop} in the $n\to 0$ limit as
\begin{equation}\label{eqn: selfloop det term1}
    \begin{aligned}
        &\lim_{n \to 0}\frac{1}{nN}\ln\mathrm{det}\left( \mY^0 \otimes \bar{\mQ}^{0} + \mY^1 \otimes \bar{\mQ}^{1} \right)\\
        =&\lim_{n \to 0}\frac{1}{nN}\ln\left(\mI_F+n\left(\widehat{p}_0 \mY^0+ \widehat{p}_1 \mY^1 \right)\left(\widehat{q}_0 \mY^0+ \widehat{q}_1 \mY^1 \right)^{-1}\right)\left(\widehat{q}_0 \mY^0+ \widehat{q}_1 \mY^1 \right)^n \\
        =&\frac{1}{N} \mathrm{Tr} \left(\widehat{p}_0 \mY^0+ \widehat{p}_1 \mY^1 \right)\left(\widehat{q}_0 \mY^0+ \widehat{q}_1 \mY^1 \right)^{-1}.
    \end{aligned}
\end{equation}
The exponential term in \eqref{APP: E selfloop} can be computed similarly, with noticing that $\mY^0$ and $\mY^1$ are rotationally invariant since $\mu=0$,
\begin{equation}\label{eqn: selfloop det term2}
    \begin{aligned}
        &\exp\bigg\{\frac{1}{2}\left( \vy^{\intercal} \mI_{\text{train}}\mX \otimes \left(\widehat{\vm}^{0}\right)^{\intercal} + \vy^{\intercal} \mI_{\text{test}}\mX \otimes \left(\widehat{\vm}^{1}\right)^{\intercal}\right)\left( \mX^{\intercal} \mI_{\text{train}}\mX \otimes \bar{\mQ}^{0}+\mX^{\intercal} \mI_{\text{test}}\mX \otimes  \bar{\mQ}^{1}\right)^{-1} \\
        & \quad \quad \times \left( \mX^{\intercal} \mI_{\text{train}}\vy \otimes\widehat{\vm}^{0}+\mX^{\intercal} \mI_{\text{test}}\vy\otimes \widehat{\vm}^{1}\right)\bigg\}\\
        =& \exp\bigg\{\mathrm{Tr}\left( n \left(\widehat{q}_0 \mY^0+ \widehat{p}_1 \mY^1 \right)\left(\widehat{m}^2_0 \mY^0+ \widehat{m}^2_1 \mY^1 \right)^{-1} \right)\bigg\},
    \end{aligned}
\end{equation}
Both \eqref{eqn: selfloop det term1} and \eqref{eqn: selfloop det term2} involve the same quantity,
\begin{equation}\label{eqn: T a b c d}
\begin{aligned}
    U(a,b,c,d) 
    &\bydef \mathbb{E}_{\mX} \left[\mathrm{Tr}\left(\left(a \mY^0+ b \mY^1 \right)\left(c \mY^0+ d \mY^1 \right)^{-1}\right) \right]\\
    &= \frac{a}{c} \mathbb{E}_{\mX}\left[ \mathrm{Tr}\left( \mY^0\left(\mY^0+\frac{d}{c} \mY^1\right)^{-1}\right)\right]+\frac{b}{d}\mathbb{E}_{\mX}\left[\mathrm{Tr}\left( \mY^1\left(\frac{c}{d}\mY^0+\mY^1\right)^{-1}\right)\right],
\end{aligned}
\end{equation}
which can be computed by random matrix free convolution \cite{voiculescu1992free}. The Green's function (also called the Cauchy function in the mathematical literature) is defined via the Stieltjes transform
\begin{equation}\label{eqn: Stieltjes}
    G_{\mY}(z):=\int \frac{\rho_{\mY}(\lambda)}{z-\lambda} \mathrm{d} \lambda =\frac{1}{N} \mathrm{Tr} \left( z \mI_N -\mY \right)^{-1},
\end{equation}
which in turn yields the spectrum transform
$
    \rho_{\mY}(\lambda)=-\left.\frac{1}{\pi} \lim _{\epsilon \rightarrow 0} \Im G_{\mY}(z)\right|_{z=\lambda+i \epsilon}.$
The corresponding Voiculescu S–transform reads
\begin{equation*}
    S_{\mY}(w)=\frac{1+w}{w} \chi(w) \quad \text{ where } \frac{1}{\chi(w)} G_{\mY}\left(\frac{1}{\chi(w)}\right)-1=w  \text{ and } \frac{1}{\chi(w)}=z.
\end{equation*}
We get the multiplicative free convolution as
\begin{equation}
    S_{\mY^0 \mY^1}(w)=S_{\mY^0}(w) S_{\mY^1}(w).
\end{equation}
After computing $S_{\mY^0(\mY^{1})^{-1}}$ and $S_{\mY^1(\mY^0)^{-1}}$, we obtain the expression for $U$ in \eqref{eqn: T a b c d}. For example, when $\tau=0.8$ and $\gamma=5$, the eigenvalue distributions of $\mY_0, \mY_1$ asymptotically follow the Marchenko--Pastur distribution as
\begin{equation*}
    \begin{aligned}
        \rho_{\mY^0}(\lambda)&=\frac{\sqrt{16-\lambda }}{8 \pi  \sqrt{\lambda }}\mathbbm{1}_{\lambda\in\{1,9\}},
        \quad \rho_{\mY^1}(\lambda)&=\frac{\sqrt{4-\lambda}}{2 \pi  \sqrt{\lambda}} \mathbbm{1}_{\lambda\in\{0,1\}}.
    \end{aligned}
\end{equation*}
We then get 
\begin{equation}
    U(a,b,c,d)=-\frac{aF}{c} \frac{-\frac{2d}{c}+\sqrt{9+\frac{16d}{c}}-3}{2 \frac{d}{c} (\frac{d}{c}-1)} -\frac{b F}{d} \frac{5-\sqrt{\frac{\frac{9c}{d}+16}{ \frac{c}{d}}}}{- \frac{2c}{d}+2}.
\end{equation}

Once $C$ and $E$ are computed, we have all the ingredients of the saddle point equation \eqref{eqn: f beta} with 12 variables, 
$$
    m_0,~m_1,~\widehat{m_0},~\widehat{m_1},~p_0,~p_1,~\widehat{p_0},~\widehat{p_1},~q_0,~q_1,~\widehat{q_0}~ \text{ and } ~\widehat{m_1}.
$$ 
A critical point of this saddle point equation gives the explicit formulas for the risks in Section \ref{APP: Derivation Sketch}.

\section{A random matrix theory approach}\label{APP: RMT}

As mentioned in the main paper, if we start with the Gaussian adjacency matrices defined before Conjecture \ref{conj: Equivalence} we can obtain some of the results described above. For simplicity, we outline this approach for the full observation case $\mI_{\text{train}} = \mI_N$, that is, for $\tau=1$, and compute the empirical risk. The partial observation case follows the same strategy but involves more complicated calculations. We let $\alpha \bydef \frac{1}{\gamma} = \frac{F}{N}$,
and rescale variables as $\sqrt{\mu\gamma}\to\mu$, $\sqrt{F}u\to u$. Following Conjecture \ref{conj: Equivalence}, we replace the binary symmetric adjacency matrix $\mA^{\text{bs}}$ by the Gaussian random matrix with a rank-one spike so that
\begin{align*}
\mA & =\boldsymbol{\Xi}^{\text{gn}}+\frac{\lambda}{N}\vy\vy^{\intercal},\\
\mX & =\boldsymbol{\Xi}^{x}+\frac{\mu}{N}\vy \vu^{\intercal}.
\end{align*}
The ridge loss reads
\begin{equation*}
    L\left(\vw\right)=\frac{1}{N}\left\Vert \vy-\mA\mX\vw\right\Vert _{2}^{2}+ \frac{r}{N}\left\Vert \vw\right\Vert^{2},
\end{equation*}
and has the unique minimum
\begin{eqnarray*}
\vw^{*} & = & \arg\min_{\vw}L\left(\vw\right)\\
 & = & \left(r\mI_F+\mX^{\intercal}\mA^{\intercal}\mA\mX\right)^{-1}\mX^{\intercal}\mA^{\intercal}\\
 & = & \left(r\mI_F+\boldsymbol{\Phi}^{\intercal}\boldsymbol{\Phi}\right)^{-1}\boldsymbol{\Phi}^{\intercal}\vy,
\end{eqnarray*}
with ${\boldsymbol{\Phi}=\mA\mX}$. We need to compute the empirical risk,
\begin{align*}
R_{\text{train}}&= \frac{1}{N}\left\Vert \vy-\mA\mX\vw^{*}\right\Vert _{2}^{2}\\
&=  \frac{r^{2}}{N}\mathrm{Tr}\left(\vy\vy^{\intercal}\mQ^{2}\right)\\
&=  -\frac{r^{2}}{N}\frac{\partial}{\partial r}\mathrm{Tr}\left(\vy\vy^{\intercal}\mQ\right),
\end{align*}
as well as the empirical loss
\begin{align}
L\left(\vw^{*}\right) & =r\vy^{\intercal}\mQ\vy/N,\label{eqn: RMT_L}
\end{align}
where we set $\mQ \bydef \left(r\mI_N+\boldsymbol{\Phi}\boldsymbol{\Phi}^{\intercal}\right)^{-1}$.

We first define four thin matrices,
\begin{equation}\label{eqn: 4 thin matrices}
\begin{aligned}
\mU=&\Biggl[&&
\frac{1}{\sqrt{N}}\boldsymbol{\Xi}^{\text{gn}}\vy+\frac{\lambda}{\sqrt{N}} \vy &\frac{\lambda}{\sqrt{N}} \vy &&\Biggr]&\in& \mathbb{R}^{N\times 2}\\
\mV=&\Biggl[&&
\frac{\mu}{\sqrt{N}}\vu & \frac{1}{\sqrt{N}}\left(\boldsymbol{\Xi}^{x}\right)^{\intercal}\vy&&\Biggr]&\in& \mathbb{R}^{F\times 2},\\
\mL=&\Biggl[&&
\mO\mV+\mU\mV^{\intercal}\mV & \mU &&\Biggr]&\in& \mathbb{R}^{N\times 4},\\
\mM=&\Biggl[&&
\mU &\mO\mV &&\Biggr]&\in& \mathbb{R}^{F\times 4}.
\end{aligned}
\end{equation}
where $\mO=\boldsymbol{\Xi}^{\text{gn}}\boldsymbol{\Xi}^x$. Then 
\begin{equation}\label{eqn: Phi}
\begin{aligned}
\boldsymbol{\Phi}&=\mU\mV^\intercal+\mO\\
\boldsymbol{\Phi}\boldsymbol{\Phi}^{\intercal}&=\mL\mM^{\intercal}+\mO\mO^{\intercal}
\end{aligned}
\end{equation}
Using Woodbury matrix identity, and denoting  $\mR=\left(r\mI_N+\mO\mO^{\intercal}\right)^{-1}$, we have
\begin{align*}
\mQ&=\left(r\mI_N+\mO\mO^{\intercal}+\mL\mM^{\intercal}\right)^{-1} \\ &=\left(r\mI_N+\mO\mO^{\intercal}\right)^{-1}-\left(r\mI_N+\mO\mO^{\intercal}\right)^{-1}\mL\left(\mI_4+\mM^{\intercal}\left(r\mI_N+\mO\mO^{\intercal}\right)^{-1}\mL\right)^{-1}\mM^{\intercal}\left(r\mI_N+\mO\mO^{\intercal}\right)^{-1}\\
 & =\mR-\mR\mL\left(\mI_N+\mM^{\intercal}\mR\mL\right)^{-1}\mM^{\intercal}\mR.
\end{align*}
Now \eqref{eqn: RMT_L} can be computed as
\begin{equation}
\frac{1}{N}\vy^{\intercal}\mQ\vy  =\underbrace{\frac{1}{N}\vy^{\intercal}\mR\vy}_{\mathbb{R}^{1\times1}}-\underbrace{\frac{1}{\sqrt{N}}\vy^{\intercal}\mR\mL}_{\mathbb{R}^{1\times4}}\underbrace{\left(\mI_4+\mM^{\intercal}\mR\mL\right)^{-1}}_{\mathbb{R}^{4\times4}}\underbrace{\frac{1}{\sqrt{N}} \mM^{\intercal}\mR\vy}_{\mathbb{R}^{4\times1}}.\label{eqn: concent 25}
 \end{equation}
The curly braces indicate $25$ random variables which all concentrate around their means (they are self-averaging in statistical physics terminology). Their expectations can be computed as follows:
\begin{itemize}
    \item $\frac{1}{N}\vy^\intercal\mR\vy$: 
    The first term of the RHS of \ref{eqn: concent 25} is a special case discussed in Section 3.2.1
    \cite{pennington2017nonlinear} and also has been discussed in \cite{dupic2014spectral}. Recalling $\mR=\left(r\mI_N+\mO\mO^{\intercal}\right)^{-1}$, we first compute the Green function \ref{eqn: Stieltjes} of $\mO\mO^{\intercal}$ as
    \[
    G_{\mO\mO^{\intercal}}\left(z\right)=\frac{1}{z}P\left(\frac{1}{z}\right)
    \]
    where $P\left(t\right)$ is the solution of
    \begin{equation}
    P=1+\frac{\left(1+\left(P-1\right)/\alpha\right)\left(1+\left(P-1\right)\right)t}{1-\left(1+\left(P-1\right)/\alpha\right)\left(1+\left(P-1\right)\right)Pt}.\label{eq: Pennington}
    \end{equation}
    Since $\mR$ is rotationally invariant, we have $$\mathbb{E}~\frac{1}{N}\vy^\intercal\mR\vy=\mathbb{E}~\frac{1}{N}\mathrm{Tr}(\vy\vy^\intercal \mR)= \mathbb{E}~\frac{1}{N} \mathrm{Tr}(\mR)=q,$$ where $q$ is the real solution of
    \[
    qr=1-\frac{\left(1+\left(qr-1\right)/\alpha\right)q}{1+\left(1+\left(qr-1\right)/\alpha\right)q}.
    \]
    It is easy to check that when $r\to0$, we have $q\to\frac{1-\alpha}{r}$ for $\alpha\leq1$.

    \item $\frac{1}{\sqrt{N}}\vy^{\intercal}\mR\mL$ and $\frac{1}{\sqrt{N}}\mM^{\intercal}\mR\vy$: we use \eqref{eqn: 4 thin matrices} and recall the definition of $\mR$ again we get
    \begin{align*}
    \mathbb{E}~\frac{1}{\sqrt{N}}\vy^{\intercal}\mR\mL =\left[\begin{array}{cc}
    \begin{array}{cc}
    \alpha\mu^{2}\lambda & \quad\end{array}\lambda & \quad\begin{array}{cc}
    \lambda & \quad\lambda\end{array}\end{array}\right]\times q,
    \end{align*}
    as well as
    \begin{align*}
    \mathbb{E}~\frac{1}{\sqrt{N}}\mM^{\intercal}\mR\vy & =\mathbb{E}\frac{1}{\sqrt{N}}\left[\begin{array}{c}
    \mU^{\intercal}\\
    \mV^{\intercal}\mO^{\intercal}
    \end{array}\right]\mR\vy=\left[\begin{array}{c}
    \lambda\\
    \lambda\\
    0\\
    0
    \end{array}\right]q.
    \end{align*}

    \item $(\mI_4+\mM^{\intercal}\mR\mL)^{-1}$: we find the entries of $\mM^{\intercal}\mR\mL$ are self-averaging, and we again use \eqref{eqn: 4 thin matrices} to average $\mM^{\intercal}\mR\mL$, 
    \begin{align*}
    \mathbb{E}~\mM^{\intercal}\mR\mL =\left[\begin{array}{cccc}
    \alpha\mu^{2}a+\alpha\lambda^{2}\mu^{2}q & \quad b+\lambda^{2}q & \quad a+\lambda^{2}q & \lambda^{2}q\\
    \alpha\lambda^{2}\mu^{2}q & \lambda^{2}q & \lambda^{2}q & \lambda^{2}q\\
    \mu^{2}c & 0 & 0 & 0\\
    \alpha\mu^{2}b & d & b & 0
    \end{array}\right],
    \end{align*}
where 
    \begin{equation}\label{eq: 5 con values}
    \begin{aligned}
    a\left(\alpha,r\right) & =\mathbb{E}~\frac{1}{N}\vy^{\intercal}\left(\boldsymbol{\Xi}^{\text{gn}}\right)^{\intercal}\mR{\boldsymbol{\Xi}^{\text{gn}}}\vy & \\
    b\left(\alpha,r\right) & =\mathbb{E}~\frac{1}{N}\vy^{\intercal}\boldsymbol{\Xi}^{x}\mO^{\intercal}\mR\boldsymbol{\Xi}^{\text{gn}}\vy & \\
    c\left(\alpha,r\right) & =\mathbb{E}~\frac{1}{N}\vu^{\intercal}\mO^{\intercal}\mR\mO\vu & \\
    d\left(\alpha,r\right) & =\mathbb{E}~\frac{1}{N}\vy^{\intercal}\mR\boldsymbol{\Xi}^x \mO^{\intercal} \mR\mO{\boldsymbol{\Xi}^{x}}^{\intercal}\vy & \\
    q\left(\alpha,r\right) & =\mathbb{E}~\frac{1}{N}\vy^{\intercal}\mR\vy.
    \end{aligned}
    \end{equation}
\end{itemize}
Now we have all the ingredients in the RHS of \eqref{eqn: concent 25}/\eqref{eqn: RMT_L}. Putting them together gives 
\begin{equation}
\begin{aligned}
 \mL\left(\vw^{*}\right) 
 &\quad=&& \frac{r}{N}\left(\vy^{\intercal}\mR\mL-\vy^{\intercal}\mR\mL\left(\mI_4+\mM^{\intercal}\mR\mL\right)^{-1}\mM^{\intercal}\mR\vy\right)\\
 &\xrightarrow{N\to \infty} && r\left(q-q^{2}\left(
\tfrac{\lambda^{2}\left(a\alpha\mu^{2}+c\mu^{2}\left(-\left(a+(b-1)^{2}\right)\right)+d\left(a\mu^{2}(c-\alpha)-1\right)+\alpha\mu^{2}+\alpha b^{2}\mu^{2}-2\alpha b\mu^{2}+1\right)}{\lambda^{2}q\left(a\alpha\mu^{2}+c\mu^{2}\left(-\left(a+(b-1)^{2}\right)\right)+d\left(a\mu^{2}(c-\alpha)-1\right)+\alpha\mu^{2}+\alpha b^{2}\mu^{2}-2\alpha b\mu^{2}+1\right)+a\mu^{2}(\alpha-c)+1}\right)\right).
\end{aligned}
\label{eq: Labcdq}
\end{equation}

The full expressions for quantities in \eqref{eq: 5 con values} are complicated. We thus analytically study the ridgeless limit $r \to 0$ in which the following hold:
\begin{equation*}
    \begin{aligned}
     a &\to \frac{\left(1-\alpha\right)^{2}}{r},\\
     b &\to \alpha-\frac{\alpha^{2}}{\left(1-\alpha\right)^{2}}r,\\
     c &\to \alpha-\frac{\alpha^{2}}{\left(1-\alpha\right)^{2}}r,\\
     d &\to 1-\frac{\alpha}{1-\alpha}r,\\
     q &\to\frac{1-\alpha}{r}+\frac{\alpha^{2}}{\left(1-\alpha\right)^{2}}.
    \end{aligned}
\end{equation*}
Substituting into \eqref{eq: Labcdq} yields
\begin{align*}
L\left(\vw^{*}\right) & \to \frac{(1-\alpha)\left(\alpha^{2}\mu^{2}+1\right)}{\alpha^{2}\left(\lambda^{2}+1\right)\mu^{2}+\alpha\lambda^{2}+1}.
\end{align*}
Finally, reverse the rescaling of $\alpha\to\frac{1}{\gamma}$, $\mu^{2}\to\mu\gamma$, we get the same expressions for $R_{\text{train}}$ as in \eqref{eqn: Risk r=0} for $\tau = 1$.

\begin{figure}[!ht]
    \centering
    \includegraphics[width=0.45\linewidth]{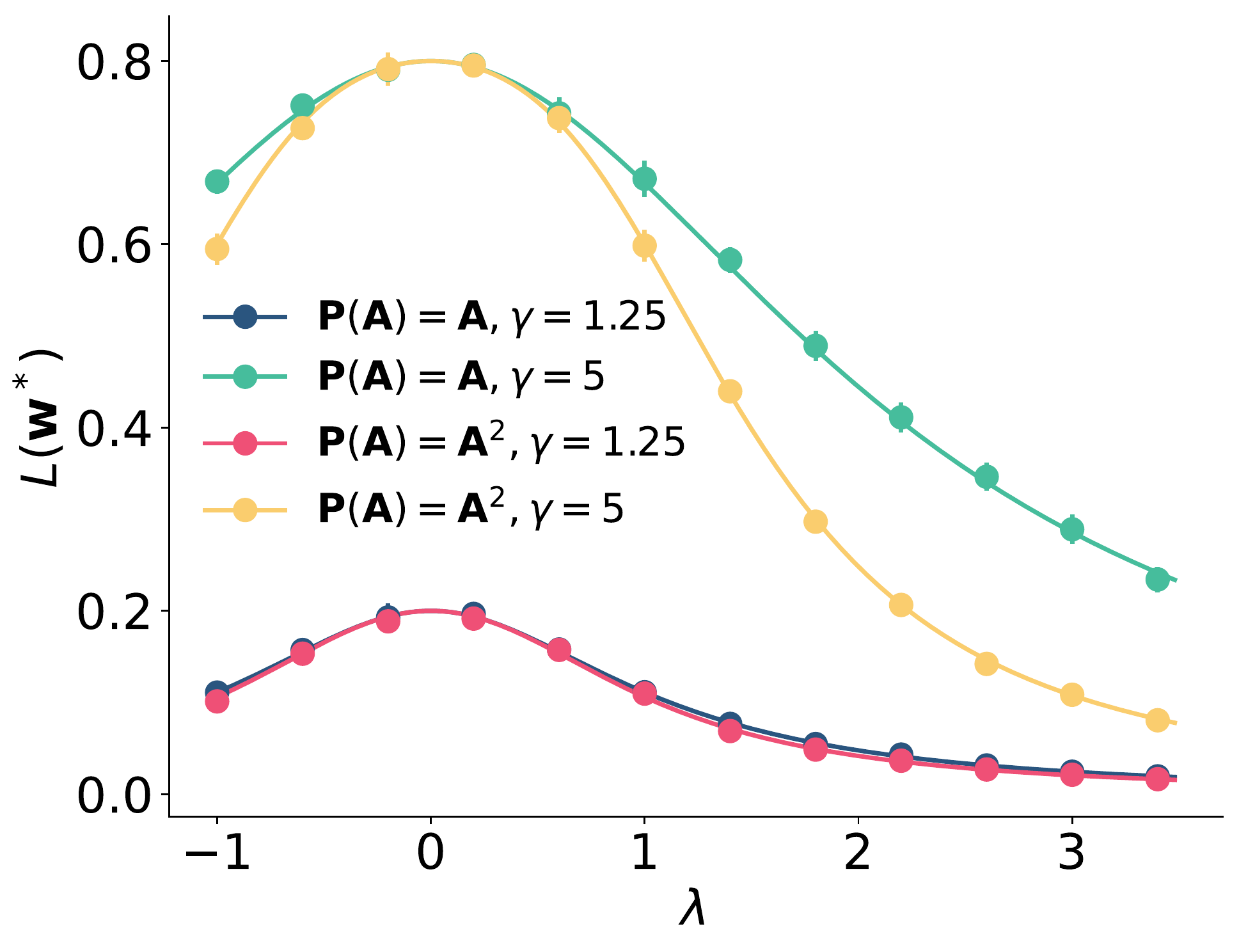}
    \caption{Theoretical results (solid line)  vs. experimental results (solid circles) for varying homophily of graphs ($\lambda$). We compare the one-hop case ($P(\mA)=\mA$) and two-hops case ($P(\mA)=\mA^2$) for non-symmetric CSBM with $\mu=0,\tau = 1, N=2000$ and $d=30$. We use non-symmetric binary adjacency matrix $\mathcal{A}^{\text{bn}}$. Each experimental data point is averaged over $10$ independent trials and the standard deviation is indicated by vertical lines.}
    \label{fig: two hops}
\end{figure}

If we assume Conjecture \ref{conj: Equivalence} and begin with Gaussian adjacency matrices, this approach can be easily extended to multi-hop by defining $\mO=\mP(\Xi^{gn})\Xi^x$ and computing corresponding $\mL,\mM,\mR$. We can then obtain a closed-form expression via \eqref{eq: Labcdq} after a longer computation. For example, when $\mP(\mA)=\mA^2$ (two hops), $\tau=1, \mu=0, r \to 0$, we get the training loss as  
\begin{equation*}
    L(\vw^*)\to\frac{(1-\alpha) \left(\alpha  \lambda ^2+1\right)}{\alpha  \left(\lambda ^2+2\right) \lambda ^2+1}.
\end{equation*}

The accurate matching between the numerical and theoretical results in Figure \ref{fig: two hops} also supports Conjecture \ref{conj: Equivalence}.

\section{A signal processing interpretation of self-loops} \label{APP: selfloops and sp}

We now show a simple interpretation of negative self-loops based on a graph signal processing intuition \cite{shuman2013emerging, ortega2018graph}. In homophilic graphs the labels change slowly on the graph: they are a low-pass signal \cite{ortega2018graph, chien2021adaptive} with most of their energy concentrated on the eigenvectors of the graph Laplacian which correspond to small eigenvalues or small ``frequencies''. Equivalently, they correspond to large eigenvalues of the adjacency matrix since $\mL = \diag(\mA \vone) - \mA$.\footnote{If node degrees are all the same the eigenvectors of the adjacency matrix and the Laplacian coincide.} On heterophilic graphs the labels usually change across an edge, which corresponds to a high-frequency signal concentrated on the small-eigenvalue eigenvectors of the adjacency matrix. A graph Fourier transform can be defined via the Laplacian but also via the adjacency matrix \cite{ortega2018graph}. The matrix product $\mA\vx=\vh$ is a delay-like filter, diagonal in the graph Fourier domain with basis functions which are the eigenvectors $\vu_1, \ldots, \vu_N$ of $\mA$. We have $(\widehat{\mA \vx})_i = \inprod{\vx, \vu_i} = \lambda_i \widehat{\vx}_i = \lambda_i \inprod{\vx, \vu_i}$, where $\lambda_i$ is the $i$-th smallest eigenvalue of $\mA$.  

\begin{figure}[h!]
    \centering
    \includegraphics[width=0.9\textwidth]{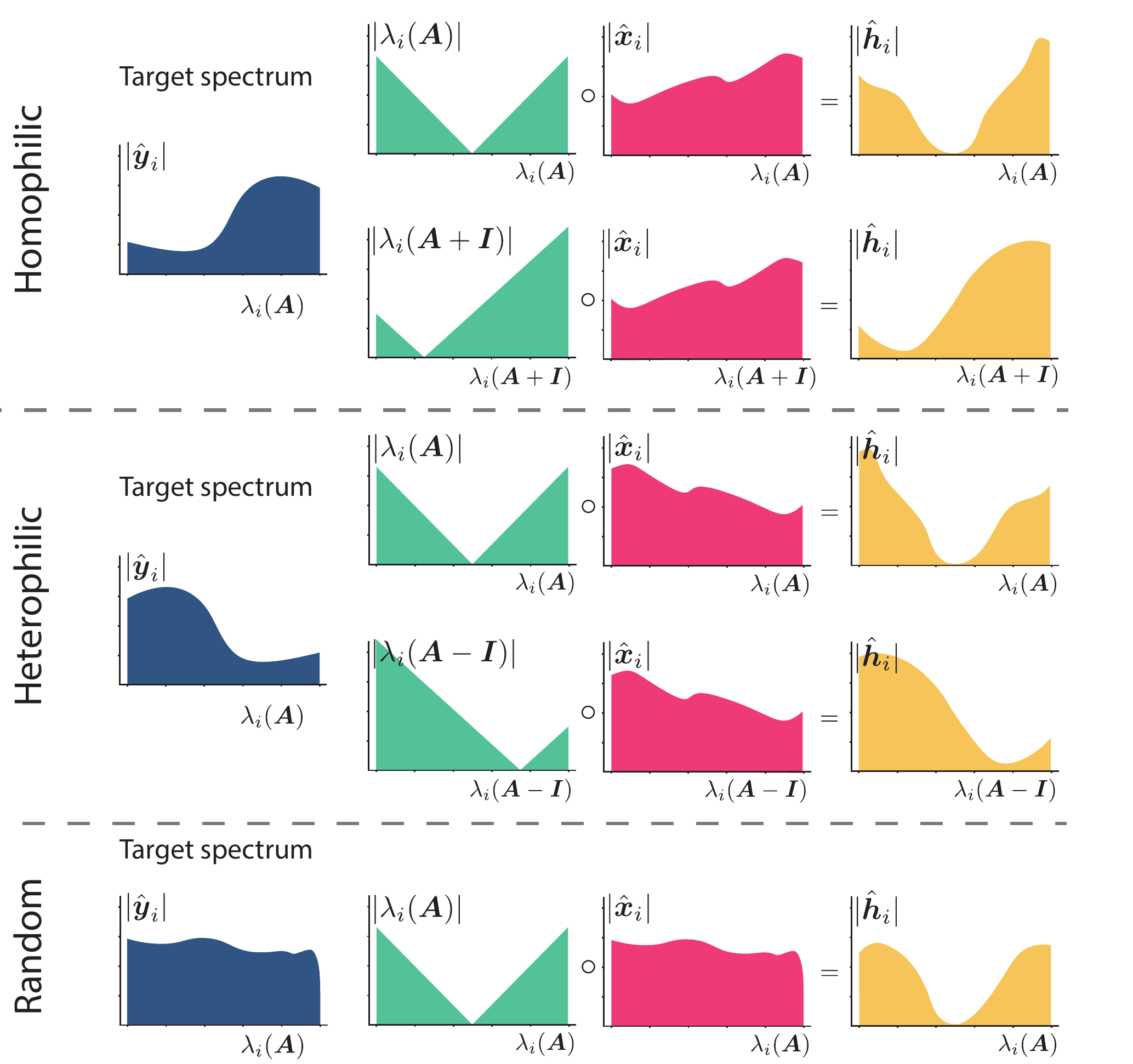}
    \caption{A spectral perspective understanding of self-loops in GCNs. The first (blue) column shows a projection of the true signal into the graph spectral domain. The following three columns illustrate the process of graph signal filtering $\mA\vx=\vh$ in the graph spectrum domain. The second column shows the eigenvalues for $\mA$ and $\mA-\mI$. The third column shows the signal $\vx$ in the spectral domain, and the forth column show the corresponding filtered signal in the spectral domain. The signal $\vx=\vy+\boldsymbol{\xi}$ is noisy, and it in general becomes closer to the target signal $\vy$ after been filtered. In the homophilic case, the signal been filtered by $\mA\vx$ is closer to the true signal compared to $(\mA+\mI)\vx$; while in the heterophilic case, $(\mA-\mI)\vx$ is better than $\mA\vx$. In all the figures, the spectral basis are arranged in the order of increasing frequency. }
    \label{fig:selfloop_ills}
\end{figure}

Figure \ref{fig:selfloop_ills} illustrates the spectra of homophilic and heterophilic labels and graphs. A homophilic graph\footnote{More precisely, a homophilic graph--label pair.} has a low-pass spectrum while a heterophilic graph has a high-pass spectrum. A self-loop shifts the spectrum of $\mA$ so that it becomes either a lowpass filter for positive $c$ or a highpass filter for negative $c$. As a result, the corresponding GCNs better suppress noise and enhance signal for the corresponding graph types. In particular, assuming that the label-related signal in $\vx$ lives between eigenvalues $\lambda_a$ and $\lambda_b$ (say, negative, so we are in a heterophilic situation), we can quantify the distortion induced by the filter $\mA + c\mI$ as ${(\lambda_a + c)}/{(\lambda_b + c)}$ which is close to 1 for large $|c|$.

\section{Double descent in various GNNs}\label{APP: moreGCN}

\begin{figure}[ht]
        \centering
        \begin{subfigure}[b]{0.475\textwidth}
            \centering
            \includegraphics[width=0.8\textwidth]{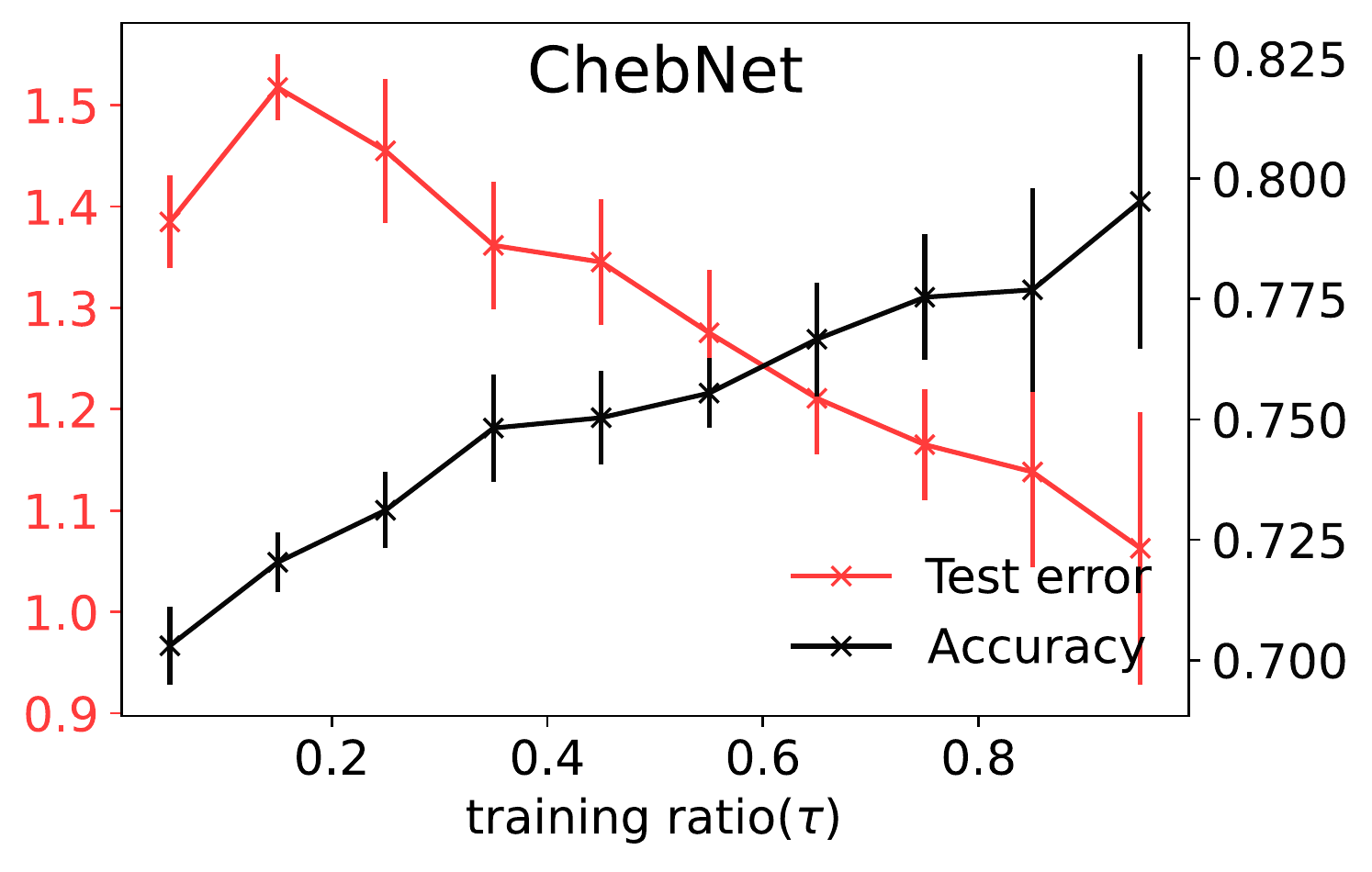}
            \caption{}\label{fig_moreGNN:a}
        \end{subfigure}
        \hfill
        \begin{subfigure}[b]{0.475\textwidth}  
            \centering 
            \includegraphics[width=0.8\textwidth]{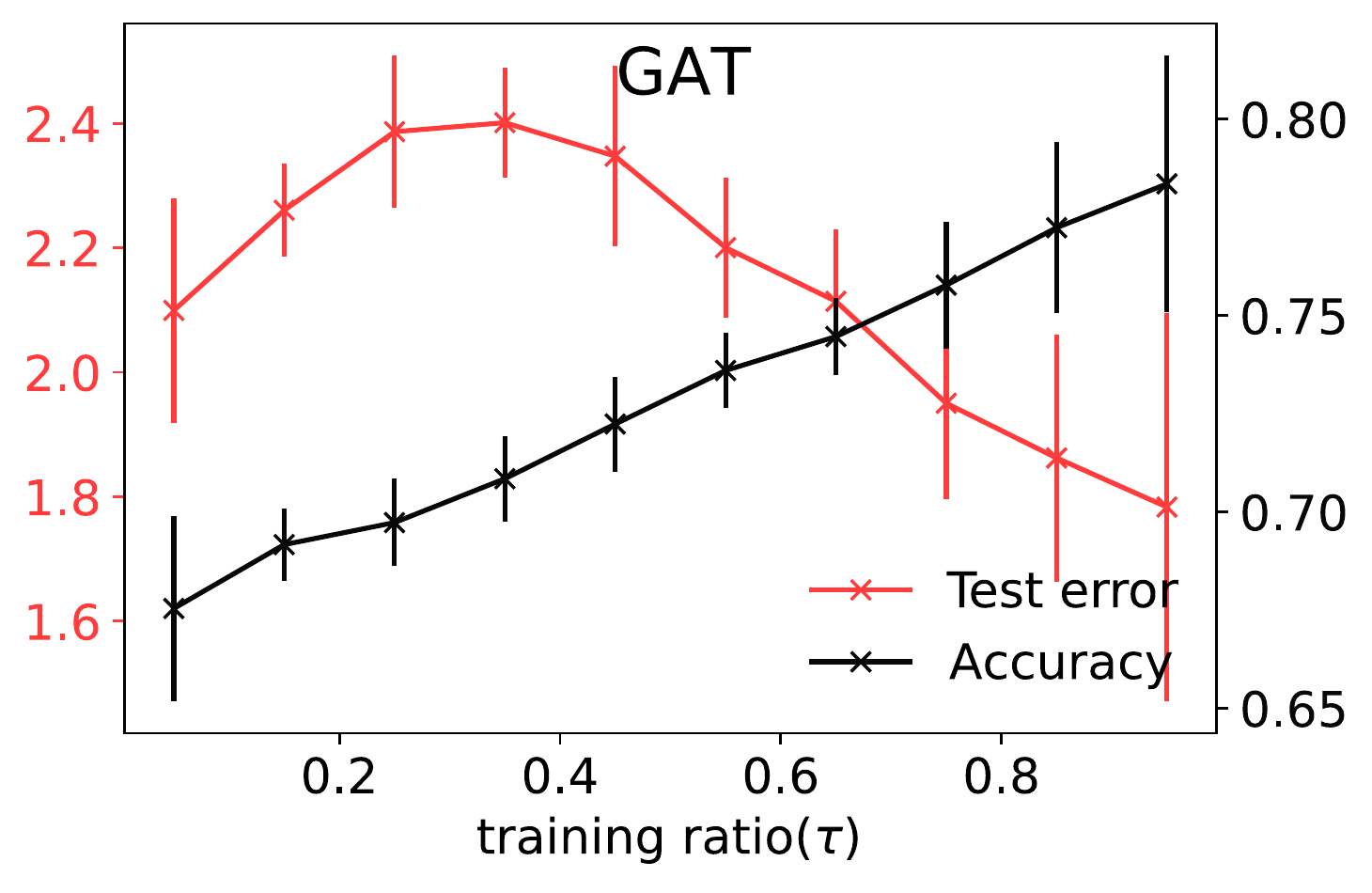}
            \caption{}\label{fig_moreGNN:b}
        \end{subfigure}

        \begin{subfigure}[b]{0.475\textwidth}  
            \centering 
            \includegraphics[width=0.8\textwidth]{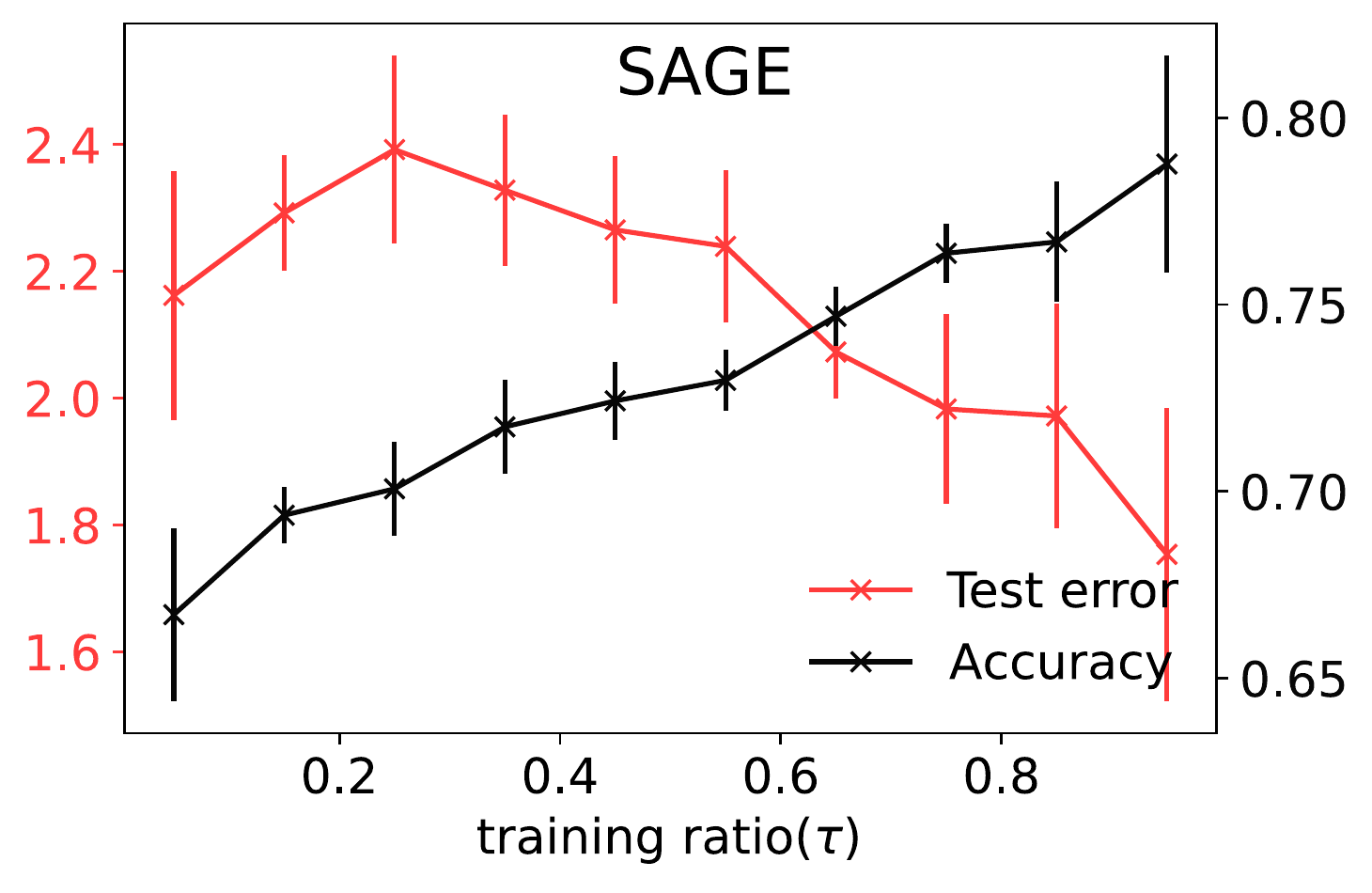}
            \caption{}\label{fig_moreGNN:c}
        \end{subfigure}
        \hfill
        \begin{subfigure}[b]{0.475\textwidth}  
            \centering 
            \includegraphics[width=0.8\textwidth]{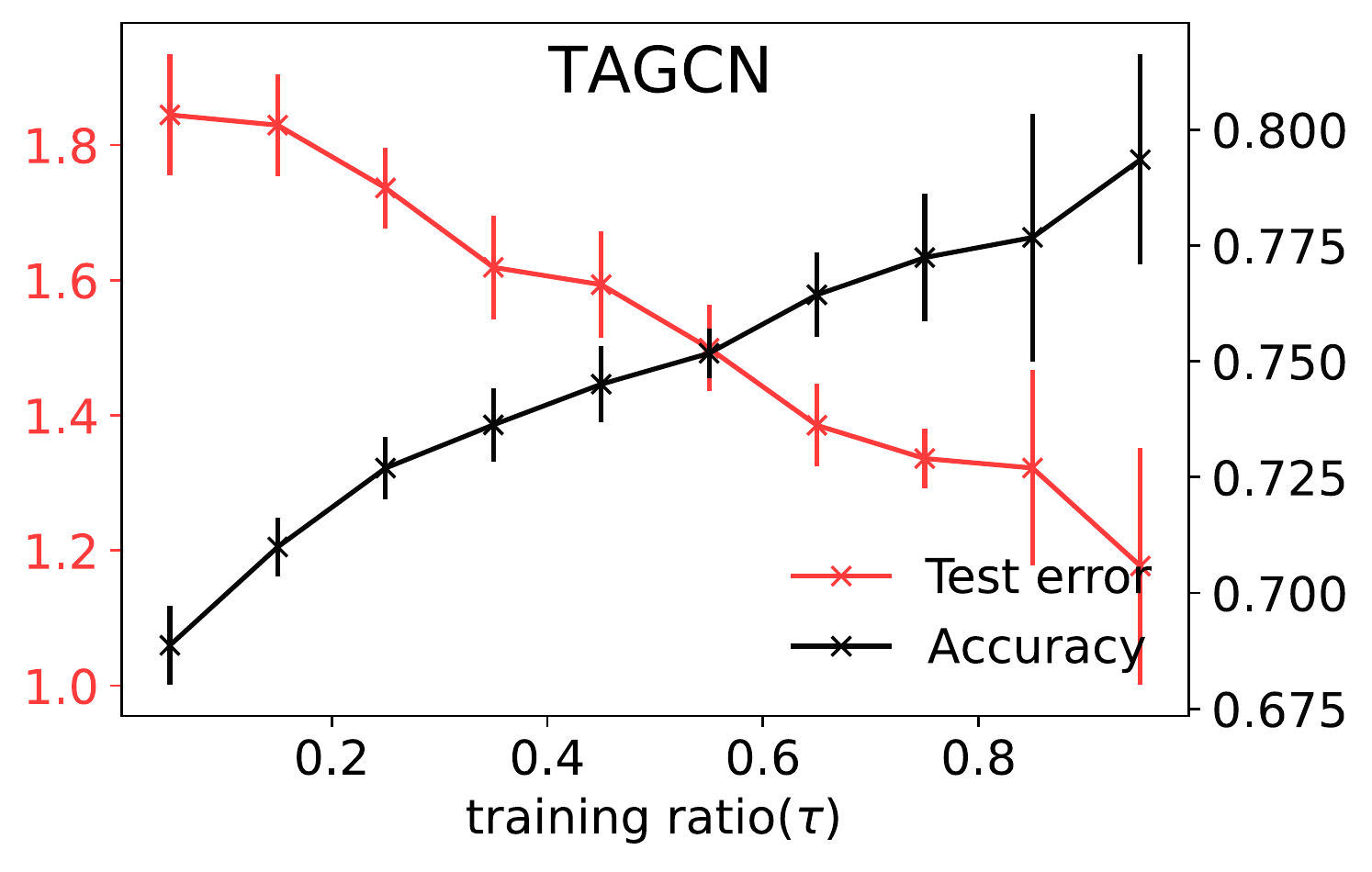}
            \caption{}\label{fig_moreGNN:d}
        \end{subfigure}
        \caption{Test error and classification accuracy at different training ratios for Chebyshev GNN (ChebNet), Graph Attention Network (GAT), the Graph Sample and Aggregate Network (SAGE), Topology Adaptive Graph Convolutional Networks (TAGCN), on the \texttt{Citeeer} dataset. All models have two layers with ReLU activations, and are trained by ADAM with the cross-entropy loss.} 
        \label{fig_moreGNN}
    \end{figure}

In Figure \ref{fig_moreGNN} we experiment with node classification on the \texttt{citeeer} dataset and some popular GNN architectures: the graph attention network \cite{velivckovic2017graph}, GraphSAGE \cite{hamilton2017inductive}, and Chebyshev graph network \cite{defferrard2016convolutional}. The architectures of these GNNs incorporate various strategies to mitigate overfitting. As a result there is no clear double descent in the test accuracy curves, but we still observe  non-monotonicity in the test risk.

\section{Experimental Details}\label{APP: experiment details}
In this section we provide more details for the experiments in the main text.  

In the real-world data experiments, all GCNs in Figure \ref{fig: Real DD} are trained by the ADAM optimizer with learning rate $10^{-2}$  and weight decay $10^{-5}$. We run ADAM for $10^4$ iterations and select the model with minimal training loss. In each trial, the training and test nodes are selected uniformly randomly. We sample training nodes separately for each label to avoid the pathology where a label has few or zero samples, which can happen at extremely low training ratios. We average $10$ different trials for each point; the error bars show their standard deviation. The standard deviation in the figures is mainly due to the train-test splits in the different trials; The fluctuations due to random initialization and stochastic optimization training are comparatively small. We do not normalize features and reprocess the data. All results in this paper are fully reproducible; code available at \url{https://github.com/DaDaCheng/SMGCN}.

\begin{table}[!h]
\centering
\begin{tabular}{c|ccccc}
Datasets                              & Cora  & Citeseer & Squirrel & Chameleon & Texas \\ \hline
Features ($F$)                       & 1433  & 3703     & 2089     & 2325      & 1703  \\
Nodes ($N$)                          & 2708  & 3327     & 5201     & 2277      & 183   \\
Edges                                & 5278  & 4552     & 198353   & 31371     & 279   \\
Inverse relative model complexity ($\gamma=N/F$) & 1.89  & 0.90     & 2.49     & 0.98      & 0.11  \\
$H(G)$                               & 0.825 & 0.718    & 0.217    & 0.247     & 0.057
\end{tabular}
\caption{Benchmark dataset properties and statistics. $H(G)$ is the level of homophily defined in \cite{pei2020geom}.}
\end{table}

For the CSBM experiments in Fig. \ref{fig_v},\ref{fig_acc} and \ref{fig_selfloop}, we calculate $\vw^*$ by \eqref{eqn: w star} and then compute \eqref{eqn: risk} and \eqref{eqn: risk}. In Fig. \ref{fig_v} and \ref{fig_acc}, we use symmetric binary adjacency matrix set $\mathcal{A}^{\text{bs}}$; In Fig. \ref{fig_selfloop} we use non-symmetric binary adjacency matrix $\mathcal{A}^{\text{bn}}$ as defined in Conjecture \ref{conj: Equivalence}. The theoretical results in Fig. \ref{fig_v}, \ref{fig: double descent with different alpha},\ref{fig_acc} and \ref{fig_selfloop} are obtained by computing the extreme values in \ref{eqn: results f}.

\end{document}